\PassOptionsToPackage{prologue,dvipsnames}{xcolor}
\documentclass[10pt,twocolumn,letterpaper]{article}

\usepackage{cvpr}              

\usepackage{graphicx}
\usepackage{tikz}
\usepackage{comment}
\usepackage{dirtytalk}
\usepackage{amsmath,amssymb} 
\usepackage{booktabs}
\usepackage{times}
\usepackage{epsfig}
\usepackage{amsfonts}
\usepackage{dsfont}
\usepackage{multirow}
\usepackage{enumitem}
\usepackage{mathtools}
\usepackage[style=plain]{floatrow}
\usepackage[normalem]{ulem}
\usepackage{balance}
\usepackage[bottom]{footmisc}
\usepackage[super]{nth}
\usepackage{subcaption}
\usepackage{cite}
\usepackage{url}
\usepackage[font=small,labelfont=bf,justification=justified]{caption}
\usepackage{tabu}
\usepackage{arydshln}

\usepackage{pifont}
\newcommand{\cmark}{\ding{51}}%
\newcommand{\xmark}{\ding{55}}%
\newcolumntype{C}[1]{>{\centering\let\newline\\\arraybackslash\hspace{0pt}}m{#1}}

\makeatletter
\newcommand{\printfnsymbol}[1]{%
	\textsuperscript{\@fnsymbol{#1}}%
}
\makeatother

\newif\ifdraft
\drafttrue

\definecolor{orange}{rgb}{1,0.5,0}
\definecolor{violet}{RGB}{70,0,170}

\ifdraft
 \newcommand{\PF}[1]{{\color{red}{\bf PF: #1}}}
 
 \newcommand{\SH}[1]{{\color{blue}{\bf SH: #1}}}
 
 \newcommand{\MT}[1]{{\color{green}{\bf MT: #1}}}

\else
 \newcommand{\PF}[1]{}
 
 \newcommand{\SH}[1]{}
 
 \newcommand{\MT}[1]{}
\fi

\newcommand{\parag}[1]{\vspace{-3mm}\paragraph{#1}}

\def\etal{\emph{et al}.}

\newcommand{\bB}{\mathbf{B}}

\newcommand{\bD}{\mathbf{D}}

\newcommand{\bE}{\mathbf{E}}
\newcommand{\bI}{\mathbf{I}}
\newcommand{\bH}{\mathbf{H}}
\newcommand{\bK}{\mathbf{K}}

\newcommand{\bPi}{\mathbf{\Pi}}

\newcommand{\bM}{\mathbf{M}}

\newcommand{\bX}{\mathbf{X}}
\newcommand{\bY}{\mathbf{Y}}

\newcommand{\mL}{\mathcal{L}}

\newcommand{\mS}{\mathcal{S}}

\newcommand{\bx}{\mathbf{x}}

\newcommand{\bp}{\mathbf{p}}
\newcommand{\bu}{\mathbf{u}}

\newcommand{\mD}{\mathcal{D}}
\newcommand{\mE}{\mathcal{E}}

\usepackage[accsupp]{axessibility}

\usepackage[dvipsnames]{xcolor}

\definecolor{cvprblue}{rgb}{0.21,0.49,0.74}
\usepackage[pagebackref,breaklinks,colorlinks,citecolor=cvprblue]{hyperref}

\def\paperID{303} 
\def\confName{3DV\xspace}
\def\confYear{2024\xspace}

\title{Unsupervised 3D Keypoint Discovery with Multi-View Geometry}

\author{Sina Honari$^{1}\thanks{work done while at Computer Vision Lab, EPFL}$
~~~~~~~Chen Zhao$^{2}$~~~~~~~Mathieu Salzmann$^{2}$~~~~~~~Pascal Fua$^{2}$
\\
{$^{1}$ Samsung AI Center Toronto} \\
{$^{2}$ Computer Vision Lab, EPFL, Switzerland}
}

\begin{document}
\maketitle

\begin{abstract}

Analyzing and training 3D body posture models depend heavily on the availability of joint labels that are commonly acquired through laborious manual annotation of body joints or via marker-based joint localization using carefully curated markers and capturing systems. However, such annotations are not always available, especially for people performing unusual activities. In this paper, we propose an algorithm that learns to discover 3D keypoints on human bodies from multiple-view images without any supervision or labels other than the constraints multiple-view geometry provides. To ensure that the discovered 3D keypoints are meaningful, they are re-projected to each view to estimate the person's mask that the model itself has initially estimated without supervision. Our approach discovers more interpretable and accurate 3D keypoints compared to other state-of-the-art unsupervised approaches on Human3.6M and MPI-INF-3DHP benchmark datasets.

\end{abstract}


\section{Introduction}

Human body joint annotations have been commonly leveraged to analyze body posture and kinematics. The 3D labels are obtained either through marker-based motion capturing~\cite{Ionescu14a, Sigal10}, wearable inertial-based tracking~\cite{lu2017whole, lopez2016wearable}, or marker-less capturing systems~\cite{Mehta17a, Joo15, Elhayek17}. These systems require considerable time, financial resources, manual labor, and sometimes specific capturing setup and clothes to obtain labels. Hence, there has been growing interest in unsupervised methods~\cite{He20c, Caron21, Bao22, He22, Rhodin19a, Honari22}, which alleviate or reduce dependence on annotations.
One way to achieve this is to discover unsupervised keypoints, distinct from pose labels, whose location and correlation are learned without any supervision by the model. 
One advantage of such a system is the discovery of meaningful 3D keypoints without any annotations. Another advantage is the reduction of the labeled annotations when the final goal is to map the discovered keypoints to the target pose of interest such as joint locations.
In this line of research, some self-supervised methods learn to find 2D keypoints ~\cite{Thewlis17a,Zhang18b, Wiles18a,Jakab18, thewlis19, lorenz19, jakab20, Schmidtke21} but do not propose a way to extend them to 3D.


\begin{figure}[htbp]
	\centering
	\includegraphics[width=0.85\textwidth]{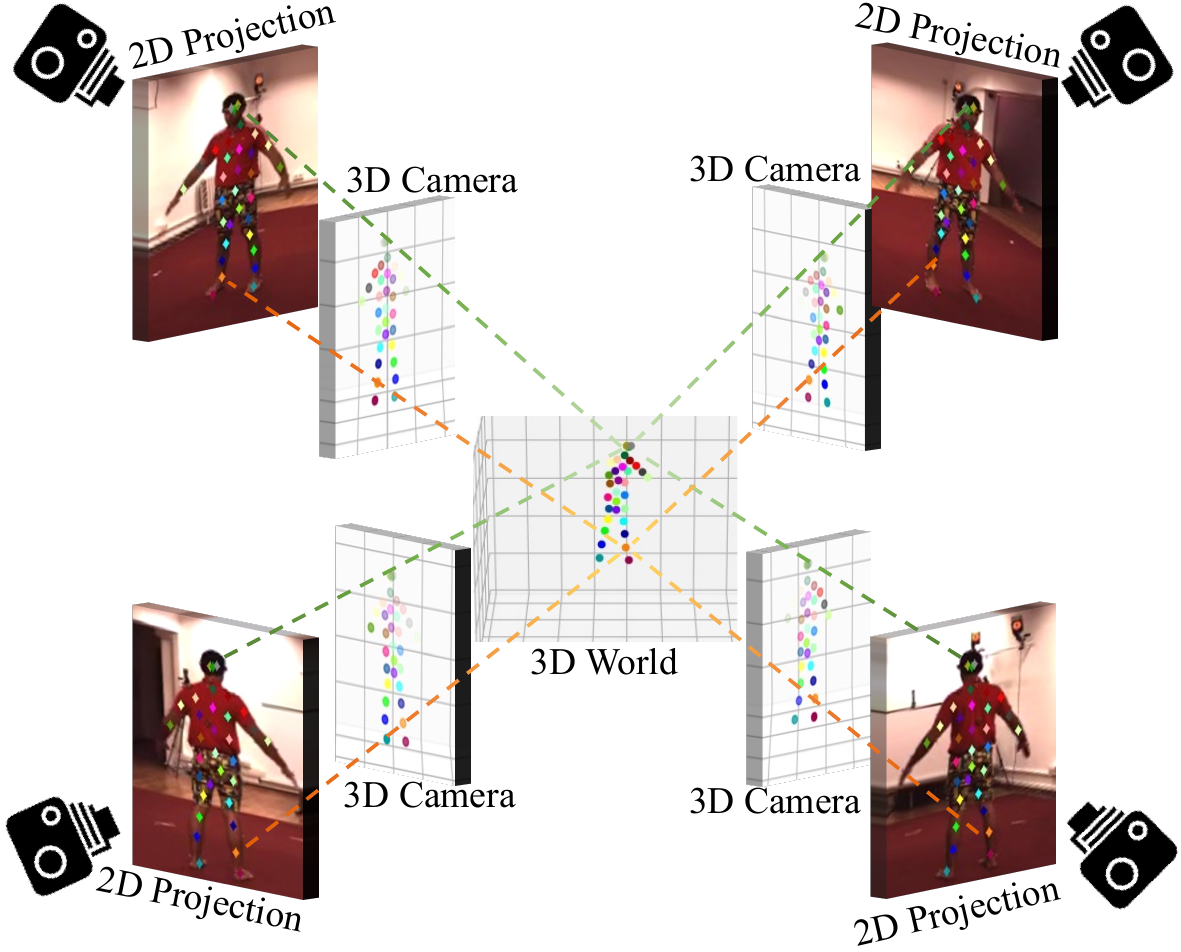}
	\caption{\small {\bf Multi-View Geometry for Unsupervised 3D Keypoint Discovery.} Our approach finds unsupervised 2D keypoints in each view and then uses multi-view geometry to construct 3D keypoints. These discovered keypoints (observed above), whose location is learned without any supervision, can be later mapped to the final pose of interest (e.g. the joint locations).}
	\label{fig:teaser}
\end{figure} 

In this work, we aim to learn in a completely unsupervised fashion to discover 3D keypoints in multi-view image setups, such as the one depicted in Fig.~\ref{fig:teaser}. We assume the cameras to be calibrated and have access to a rough estimate of the background image of the scene but do not require 2D or 3D annotations, mask labels, or pre-trained models. Some approaches~\cite{Suwajanakorn18a, Noguchi22, Chen21f} also extract unsupervised 3D keypoints from images. However, they have only been demonstrated on rigid toy datasets with controllable dynamics and limited variations. By contrast, we aim at real datasets, featuring people performing diverse activities.

\begin{figure*}[htbp]
	\centering
	\includegraphics[width=0.9\textwidth]{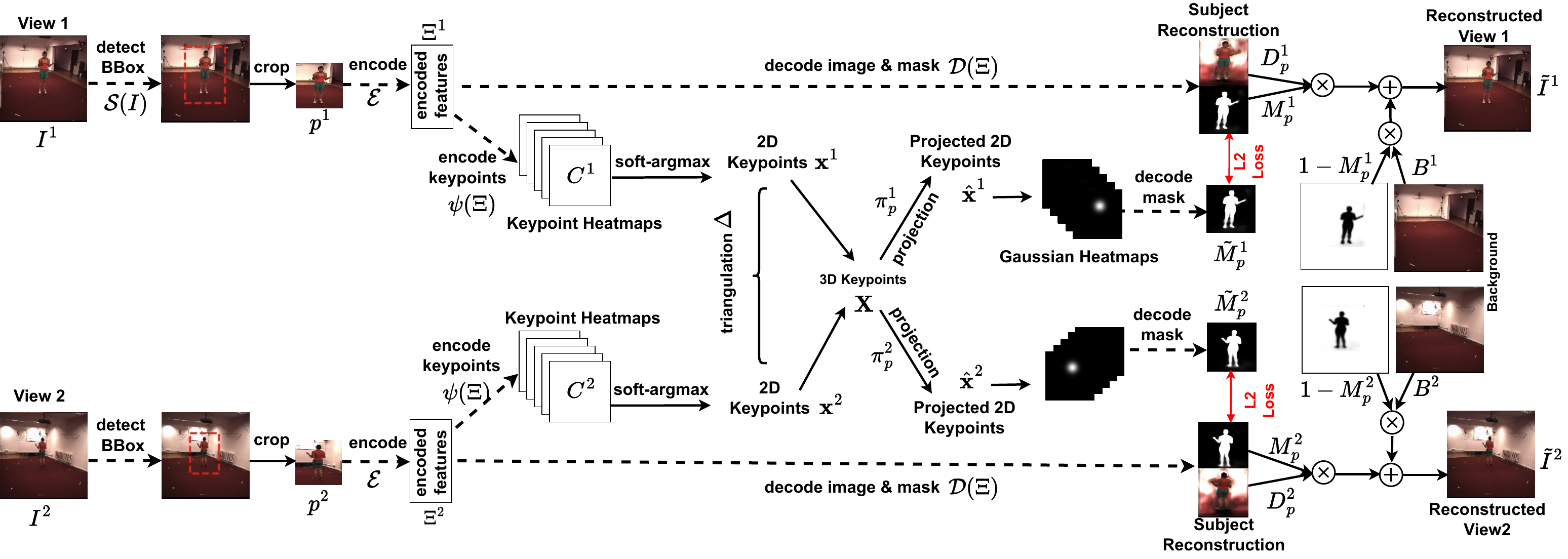}
	\caption{\small {\bf Approach.} Given images from different views and an estimate of the background image, the model first detects and crops the subject in each view. The cropped patch is then passed to an encoder that encodes only the foreground subject information $\Xi$ through reconstruction of the input image by prediction of the foreground mask $\bM_{\bp}$. This constitutes the image-reconstruction path.
	The encoded features $\Xi$ are then used to discover 2D keypoints $\bx$ by applying a 2D soft-argmax to each keypoint channel. The 2D keypoints from different views are then triangulated to obtain 3D keypoints $\bX$ in the world-coordinate using full camera projection matrices, which are then projected separately to each view to obtain the view-specific 2D keypoints $\hat{\bx}$. These 2D keypoints are then used to construct a mask $\tilde{\bM}_{\bp}$ by minimizing its difference to the mask $\bM_{\bp}$ predicted by the model itself in the image-reconstruction path. No label is used for subject detection, mask reconstruction, or keypoint estimation. The dashed line indicate trainable model components.}
	\label{fig:diagram}
\end{figure*}
In this paper, given a multi-view setup, our model predicts the masks of the foreground subject in each view in a self-supervised way. In parallel, the model discovers unsupervised 2D points from each camera's point of view. These 2D detections are triangulated into potential 3D keypoints. The potential 3D keypoints are then re-projected onto maps in each view to reconstruct the subject's mask that the model has itself predicted. \textbf{The training is completely unsupervised. No labels such as 2D or 3D annotations or masks are used during training.}

Our complete pipeline is depicted by Figure~\ref{fig:diagram}. This framework guarantees that the discovered 3D keypoints satisfy multi-view projection properties while correlating strongly with the 3D posture of the subject. The 3D keypoints can then easily be mapped to the target 3D pose using a simple linear or a multi-layer perceptron network.  
In short, we propose a real-world unsupervised 3D keypoints discovery method that can be applied directly to uncurated images.
We show that our approach outperforms state-of-the-art unsupervised learning methods on Human3.6M and MPI-INF-3DHP benchmark datasets.

\section{Related Work}
\label{sec:related}


The objective of unsupervised learning models \cite{He20c, Caron21, Bao22, He22,Rhodin19a, Honari22} is to pretrain neural networks to be later fine-tuned on the tasks of interest or to train models to extract salient features. These features are then used for downstream applications with a small amount of data using a simple mapping, such as linear adaptation \cite{He20c, Caron21, Bao22, He22}, or a two hidden layer multi-layer perceptron (MLP) \cite{Rhodin18b, Rhodin19a, Honari22} among others. In this section, we briefly review unsupervised models for keypoint detection and pose estimation. In this paper, we make a distinction between \textit{keypoint}, found by a model in an unsupervised manner, and \textit{pose}, corresponding to dataset labels such as joint locations.

\textbf{Unsupervised 2D Keypoint Discovery.} 
There have been multiple approaches ~\cite{Thewlis17a,Zhang18b, Wiles18a,Jakab18, thewlis19, lorenz19, jakab20, Schmidtke21} that propose unsupervised 2D keypoint discovery. However, the extension of these models to 3D keypoint estimation has not been explored. Moreover, these methods usually rely on center-cropped subject images, while our approach can directly work on uncurated images. 

\textbf{Unsupervised 3D Pose Models.} 
On the other hand, some approaches learn unsupervised 3D poses from 2D poses using cross-views re-projection and matching \cite{Kocabas19, Chen19f, Iqbal20, Li20i, Wandt21, Rhodin18a}, or temporal consistency \cite{Tripathi20, Yu21c, Hu21c}. Some papers instead learn a prior over the 2D pose distribution and enforce consistency across single frame lifting, rotation, and projection \cite{Chen19g, Wandt22}. However, all of these approaches rely on either 2D labels or pre-trained 2D pose estimation models, which still depend on a considerable number of annotations. In this paper, we aim at finding unsupervised keypoints directly from images, without relying on any labels or pre-trained models. 

\textbf{Unsupervised Pose-Relevant Features.}
The works in \cite{Denton17, Rhodin18b, Rhodin19a, Honari22} apply unsupervised learning directly to input images to extract pose-relevant information. In particular, the approaches in \cite{Denton17, Honari22} apply temporal learning to extract time-variant information, while the approaches in \cite{Rhodin18b, Rhodin19a} leverage multi-view geometry by applying latent features rotation from one view to another. These latent features cannot be visualized as keypoints, as they live in uninterpretable latent layers. Contrary to these approaches, we apply a full-projection matrix including both intrinsics and the translation component of the extrinsics. This allows directly projecting the estimated 3D keypoints back to the images using complete projection matrices, which contributes to further structure into the latent space, hence yielding interpretable 3D keypoints that can be visualized and correlate strongly with the subject's pose. In practice, this intrinsic structure contributes to higher pose accuracy.

Kundu \etal~\cite{Kundu20a, Kundu20b} also propose unsupervised 3D pose estimation models through pose and shape disentanglement and a spatial transformation map~\cite{Kundu20b} or a part-based body model \cite{Kundu20a}. The proposed framework requires access to a known shape model and body kinematic priors. Contrary to these approaches, we do not resort to such priors and learn directly from images with a much simpler pipeline.

\textbf{Unsupervised 3D Keypoint Discovery.}
Some approaches extract unsupervised 3D keypoints from 3D point clouds \cite{Fernandez20, Chen20e} or a 3D shape \cite{Jakab21}. Contrary to these approaches, we aim at extracting unsupervised 3D keypoints from only input images. On the other hand, the approaches in \cite{Suwajanakorn18a, Noguchi22, Chen21f} learn 3D keypoints from input images using a multi-view learning framework. In particular, Chen \etal~ \cite{Chen21f} learn latent 3D keypoints for control in reinforcement-learning problems. Noguchi \etal~\cite{Noguchi22} estimate keypoints by applying a volumetric rendering using signed-distance-functions to reconstruct all views. This creates a complex setup that requires at least 5 cameras for training. Our model, on the other hand, is much simpler to train, with fewer loss terms, and can even train with 2 cameras.

Closest to our approach is \cite{Suwajanakorn18a}, which learns unsupervised 3D keypoints on rigid objects. While this approach also leverages multi-view geometry, similar to \cite{Noguchi22, Chen21f} it has been only evaluated on toy datasets, with the possibility of rendering many views and limited diversity in appearance, pose, and depth values. As we will show later, this approach is prone to limitations when dealing with real-world data. Moreover, none of these approaches handle foreground subject extraction, as they use toy datasets with no or simple backgrounds, with \cite{Suwajanakorn18a, Noguchi22} requiring ground-truth masks. Sun \etal~\cite{Sun23bkind} also propose an unsupervised spatio-temporal multi-view keypoint discovery for real-world data. The proposed model builds a 3D volume and applies multiple projections and re-projection in-between 2D and 3D spaces before constructing an edge graph on keypoints to constrain the joint lengths. Our approach has a much simpler pipeline and does not require temporal information to discover its keypoints, while obtaining higher accuracy in practice.


\section{Method}
\label{sec:method}

Our goal is to train a network in an unsupervised fashion to discover 3D keypoints in multi-view calibrated images.  To ensure the keypoints are meaningful, we need to guarantee they encode information about only the foreground subject rather than the background. To this end, the model should first detect the subject and then encode the foreground information. This is achieved by reconstructing the input image through the prediction of a mask that separates the foreground from the background, hence allowing to encode only the foreground information.

Our network first encodes the foreground subject and uses the resulting image features to discover potential 2D keypoints in each view. The 2D keypoints from different views are then triangulated using calibration data to create potential 3D keypoints expressed in world coordinates. These 3D keypoints are the features of interest in our model. Without any backward loop, there is no guarantee that they are meaningful since there are no keypoint labels. Hence, they are re-projected to each view to obtain 2D view-dependent keypoints, which are then passed to a mask decoder that reconstructs the foreground subject mask. This process is depicted in Figure~\ref{fig:diagram}. In the following sections, we describe the pipeline components in more detail.




\subsection{View-Dependent Feature Extraction}
\label{sec:view-dependent}

In this section, we describe how detection, encoding, and decoding components are used to extract view-dependent information and the foreground mask. Given an input image $\bI$, a spatial transformer network (STN)~\cite{Jaderberg15}  $\mS$ is used to extract four parameters; two of which specify the scale $s^x, s^y$ and the other two the center of the bounding box $u^x, u^y$, yielding $\mS(\bI)=(s^x, s^y, u^x, u^y)$. A patch $\bp$ is then cropped using the bounding box coordinates and is then passed to an encoder $\mE$ yielding $\mE({\bp}) = {\Xi}$.




A decoder $\mD$ then takes the encoded features and outputs an RGB image patch $\tilde{\bD}_{\bp}$ together with a foreground mask patch $\bM_{\bp}$, which can be written as
\begin{align}
\mD({\Xi}) =
(\bD_{\bp}, \bM_{\bp}). 
\label{eq:decoder}
\end{align}

Finally, the inverse operation of the spatial transformer network (STN) is applied to put the patches back into the full-image resolution, yielding ${\mS}^{-1}(\bD_{\bp}, \bM_{\bp}) = (\bD, \bM)$. Decoded image $\bD$ is then merged with the background image $\bB$\footnote{The background image is obtained by taking a per-pixel median of every frame in the video sequence.}, using the predicted mask $\bM$ to reconstruct the input image $\tilde{\bI}$. This operation is equal to 
\begin{align}
\tilde{\bI}=   \bM \times \bD + (1- \bM) \times \bB \;,
\label{eq:synthesize}
\end{align}

where $\times$ indicates Hadamard product.
This process is depicted in Figure \ref{fig:diagram}. The input image reconstruction is crucial to capture information 
only about the foreground subject into the latent features $\Xi$ in addition to estimating the foreground subject mask $\bM_{\bp}$. Note that all components including foreground subject detection and mask prediction are unsupervised. No bounding box or mask labels are used. They are all trained through input image reconstruction.

\subsection{Unsupervised 3D Keypoint Discovery}
\label{sec:Unsup-3D-kpts}

The latent information in each view ${\Xi}^{\vartheta}, \vartheta \in \{1,\dots, V\}$, with $V$ being the total number of views, is passed to a 2D keypoint encoder $\psi$ that outputs $N$ channels, one for each potential keypoint $n$. This yields
\begin{align}
\psi({\Xi}) = \{C_n : n = 1,\dots, N\}, 
\end{align}
where we refer to the set $\{C_n\}_{n=1}^{N}$ by $C$. Soft-argmax operation~\cite{Honari18} is then applied to each channel $C_n$ to obtain a potential 2D keypoint $\bx_n = (u, v)_{n}$. This can be written as
\begin{align}
\begin{bmatrix} u_n \\ v_n \end{bmatrix} =
\sum_{i,j} \text{softmax}(C_n)_{i,j} \begin{bmatrix} i \\ j \end{bmatrix} \;,
\label{eq:softargmax}
\end{align}
where the standard softmax operation is applied to each channel $C_n$ to obtain a probability distribution, with $i, j$ indicating the row and column coordinates of the channel. In summary, this process obtains a weighted average of the coordinates, where the weights are given by the softmax probability map.

Given an intrinsic matrix $\bK_\bI$ specifying the projection from the camera coordinates to the pixel coordinates on image $\bI$ and an extrinsic matrix $\bE$ giving the change of coordinate system from camera to world coordinates, the camera projection matrix on the full image resolution can be written as $ \bPi_\bI = \bK_\bI \bE$. 

Having the scale parameter $(s^x, s^y)$ and the top-left coordinates $({b}^x, {b}^y)$ of the bounding box estimated by the spatial transformer network $\mS$ in Section~\ref{sec:view-dependent}, the intrinsic matrix of the full-resolution image
\begin{align} 
\bK_\bI = 
\begin{bmatrix}
f^x & 0 & c^x\\
0 & f^y & c^y\\
0 & 0 & 1
\end{bmatrix}
\end{align}
is updated to correspond to the detected patch $\bp$ by using
\begin{align} 
\bK_{\bp} = 
\begin{bmatrix}
s^x f^x & 0 & s^x (c^x - {b}^x) \\
0 & s^y f^y & s^y (c^y - {b}^y) \\
0 & 0 & 1
\end{bmatrix}.
\end{align}
This yields the updated projection matrix of the detected patch $\bPi_{\bp}= \bK_{\bp} \bE$.

For each keypoint $n \in \{1,\dots, N\}$, its corresponding 2D locations $\{ \bx_n^\vartheta \}_{\vartheta = 1}^{V}$ and projection matrices $\{\bPi_{\bp}^\vartheta \}_{\vartheta = 1}^{V}$ from all $V$ views are then passed to a triangulation operation $\Delta$ to output potential 3D keypoint $\bX_n=(x,y,z)_n$ by
\begin{align}
\Delta(\{ \bx_n^\vartheta \}_{\vartheta = 1}^{V}, \{\bPi_{\bp}^\vartheta \}_{\vartheta = 1}^{V}) = \bX_n \;.
\label{eq:triangulation}
\end{align}
We use direct linear transform~\cite{Hartley00} for triangulation, which is differentiable and allows back-propagation through the keypoints. Note that this triangulation operation is applied separately for each keypoint. The semantic consistency of the keypoints across all views is enforced through multi-view geometry, as the $n$-th keypoint from different views are mapped together.

The potential 3D keypoints $\{ \bX_n \}_{n=1}^{N}$ are the features of interest that we will eventually use from the network's prediction. The obtained keypoints so far are not guaranteed to contain any useful information as they are unsupervised. 
We project them to the foreground subject mask to constrain the keypoints in order to make them correspond to the subject of interest. To do so, each 3D keypoint $\bX_n$ is first re-projected to each view $\vartheta$ by using $\bPi_{\bp}^{\vartheta}$, this yields
\begin{align}
\hat{\bx}_n^{\vartheta} = (\hat{u}, \hat{v})_{n}^{\vartheta} = \bPi_{\bp}^{\vartheta}(\bX_n) \;.
\label{eq:reproject}
\end{align}

The set of re-projected 2D keypoints $\{\hat{\bx}_n^{\vartheta}\} _{n=1}^{N}$ in each view is then passed to a mask-decoder $\phi$ that estimates the mask of the foreground subject $\tilde{\bM}_{\bp}^{\vartheta}$, which is equivalent to
\begin{align}
\tilde{\bM}_{\bp}^{\vartheta} = \phi(\{\hat{\bx}_n^{\vartheta}\} _{n=1}^{N}) \;.
\label{eq:mask_reconstruct}
\end{align}

This creates a full pipeline validating the discovered 3D keypoints that correspond to the foreground subject and satisfy the multi-view constraints. The semantic information of keypoints is learned automatically by the model without supervision. 

\subsection{Unsupervised Training Losses}
\label{sec:unsup_losses}

To train the unsupervised keypoint discovery model we use the following losses. We drop the view-dependent notations $\vartheta$, as the following losses apply to all views.

\textbf{Image Reconstruction Loss.} 
The reconstructed image $\tilde{\bI}$, described in Section~\ref{sec:view-dependent}, should match the input image ${\bI}$ using two losses; an RGB pixel-wise loss, and a perceptual loss using ImageNet features~\cite{Johnson16b, Rhodin19a, Honari22}; comparing the extracted features of the first 3 layers of the ResNet18~\cite{He16a}. This can be written as
\begin{align}
\mL_{\text{reconst}} =  {\| \bI - \tilde{\bI}\|}_2^2 + \beta \sum_{l=1}^{3}{\| \text{Res}_{l}(\bI) - \text{Res}_{l}(\tilde{\bI})\|}_2^2 \;.
\label{eq:loss_reconst}
\end{align}
%

\textbf{Mask Reconstruction Loss.} 
The estimated mask patch $\bM_{\bp}$ in the image-reconstruction module of Section~\ref{sec:view-dependent}, is taken as target for the mask $\tilde{\bM}_{\bp}$ predicted from discovered keypoints in Section~\ref{sec:Unsup-3D-kpts} to minimize
\begin{align}
\mL_{\text{mask}} = {\| \tilde{\bM}_{\bp} - \bM_{\bp}\|}_2^2 \;.
\label{eq:loss_mask}
\end{align}

In order to further enforce the predicted keypoints lie in the foreground mask $\bM_{\bp}$ and enhance the location of the detected bounding box, the two following losses are also used. 

\textbf{Coverage Loss.} 
 Each projected keypoint $\hat{\bx}_n$ should be only on the foreground subject mask $\bM_{\bp}$ and not the background. To enforce this, a Gaussian heatmap $\bH_n$ is first generated by taking its mean to be the keypoint $\hat{\bx}_n$~\footnote{its standard deviation was set to a fixed small value, $\sigma=0.02$ in our experiments.}. This map is then normalized to have a probability map that sums to one, which equals to
\vspace{-0.25cm}
\begin{align}
\bar\bH_n =  \bH_n / \sum_{i,j} (\bH_n^{i,j}) \;.
\label{eq:prob_keypoint_hmap}
\end{align}

The coverage loss in Eq.~(\ref{eq:loss_coverage}) then maximizes the linear projection of $\bar\bH_n$ onto $\bM_{\bp}$ for each keypoint $\hat{\bx}_n$. This ensures the probability mass in $\bar\bH_n$ falls into locations of the mask $\bM_{\bp}$ that correspond to the foreground subject. This can be written as 
\vspace{-0.25cm}
\begin{align}
\mL_{\text{coverage}} =  \frac{1}{n} \sum_{n=1}^{N}{{| 1 -  \bar\bH_n \odot \bM_{\bp} |}} \;,
\label{eq:loss_coverage}
\end{align}
where $\odot$ indicates the dot product. 

\textbf{Bounding Box Centering Loss.} 
To ensure the detected bounding box completely covers the foreground subject, we enforce the center of the predicted keypoints on the subject to be equal to the center of the detected bounding box. 
Considering $\bu = (u^x, u^y)$ to be the center of the bounding box predicted by $\mS$, as described in Section~\ref{sec:view-dependent}, and the set of re-projected keypoints on the subject to be $\{\hat{\bx}_n\}_{n = 1}^{N}$, the loss then equals to  
\vspace{-0.4cm}
\begin{align}
\mL_{\text{centering}} =  {| \bu -   \frac{1}{N} \sum_{n=1}^{N}\hat{\bx}_n |} \;.
\label{eq:loss_bbox_center}
\end{align}

This loss together with $\mL_{\text{reconst}}$ of Eq.~(\ref{eq:loss_reconst}) force the predicted bounding box to completely cover the keypoints and hence completely detect the subject.
The total loss for training the model is then equal to

\vspace{-0.5cm}
\begin{align}
\mL_{\text{unsup}} =  \mL_{\text{reconst}} + \gamma \mL_{\text{mask}} + \delta \mL_{\text{coverage}} + \eta \mL_{\text{centering}} \;.
\label{eq:loss_unsup}
\end{align}
%

\subsection{Single-View Keypoint Estimation}
\label{sec:unsup_sv}
The procedure explained in Sections \ref{sec:Unsup-3D-kpts} and \ref{sec:unsup_losses} trains an unsupervised multi-view keypoint discovery model. Once this network is trained, one can take the discovered multi-view 3D keypoints $\bX = \{\bX_n\}_{n=1}^{N}$ as targets to train an unsupervised single-view 3D keypoint estimation model. Since the 2D keypoints $\bx = \{\bx_n\}_{n=1}^{N}$, are found in each view independently, it is only needed to lift them to 3D. Hence, we train a simple residual MLP model \cite{Martinez17a}, which maps $\bx$ to the set of discovered 3D keypoint targets $\bX$. The single-view 3D lifting network $\psi$ is trained using mean-squared error between the predicted and target 3D keypoints:
\vspace{-0.25cm}
\begin{align} 
\mL_{\text{SV}} =  {\|\psi(\bx) - \bX \|}_2^2 \;.
\label{eq:loss_sv}
\end{align}

\subsection{Pose Estimation}
\label{sec:pose_estimation}
Once the unsupervised model is trained, 3D keypoints are extracted using the procedures explained in Sections~\ref{sec:Unsup-3D-kpts} and \ref{sec:unsup_sv}, for multi-view and single-view setups respectively. To verify the quality of the discovered unsupervised 3D keypoints $\bX = \{\bX_n\}_{n=1}^{N} $, we map them to the pose labels $\bY = \{ \bY_m\}_{m=1}^{M}$, specifying body-joint locations, using a simple linear or two hidden-layer multi-layer perceptron (MLP) $\theta$ by minimizing
\begin{align} 
\mL_{\text{pose}} =  {\| \theta(\bX) - \bY \|}_2^2 \;.
\label{eq:loss_pose3d}
\end{align}

Note that the pose labels are only used to evaluate the quality of the discovered 3D keypoints through this mapping. They are not used for training the unsupervised 3D keypoint model, described in Sections \ref{sec:view-dependent} to \ref{sec:unsup_sv}.


\begin{figure*}[h!]
	\centering
        \begin{subfigure}[b]{0.45\textwidth}
	\includegraphics[width=\textwidth]    {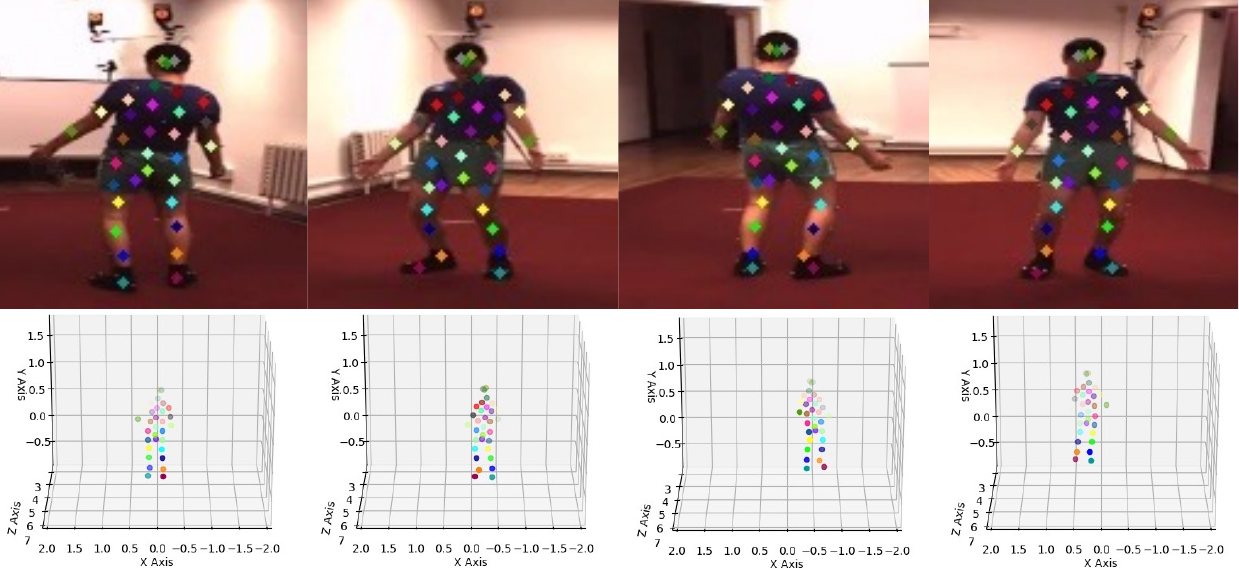}
	\end{subfigure}
	\hspace{0.5cm}
	\begin{subfigure}[b]{0.45\textwidth}
		\includegraphics[width=\textwidth]{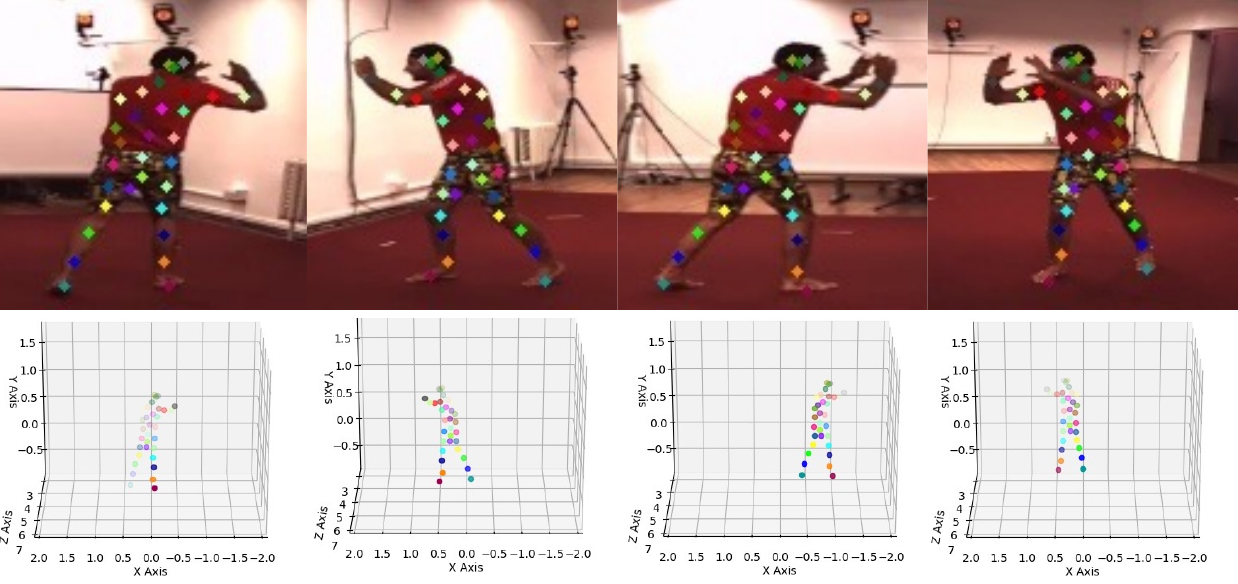}
	\end{subfigure}
	\vspace{-2mm}
	\caption{\small 2D and 3D keypoints found by a 32-keypoint prediction model on H36M. The 2D keypoints are consistent across views and the 3D keypoints capture the posture of the person, which indicates that they correlate with the person's pose.}
	\label{fig:kpts_3D_H36M}
\end{figure*}

\begin{figure*}[h!]
	\centering
	\begin{subfigure}[b]{0.45\textwidth}
		\centering
		\includegraphics[width=\textwidth]{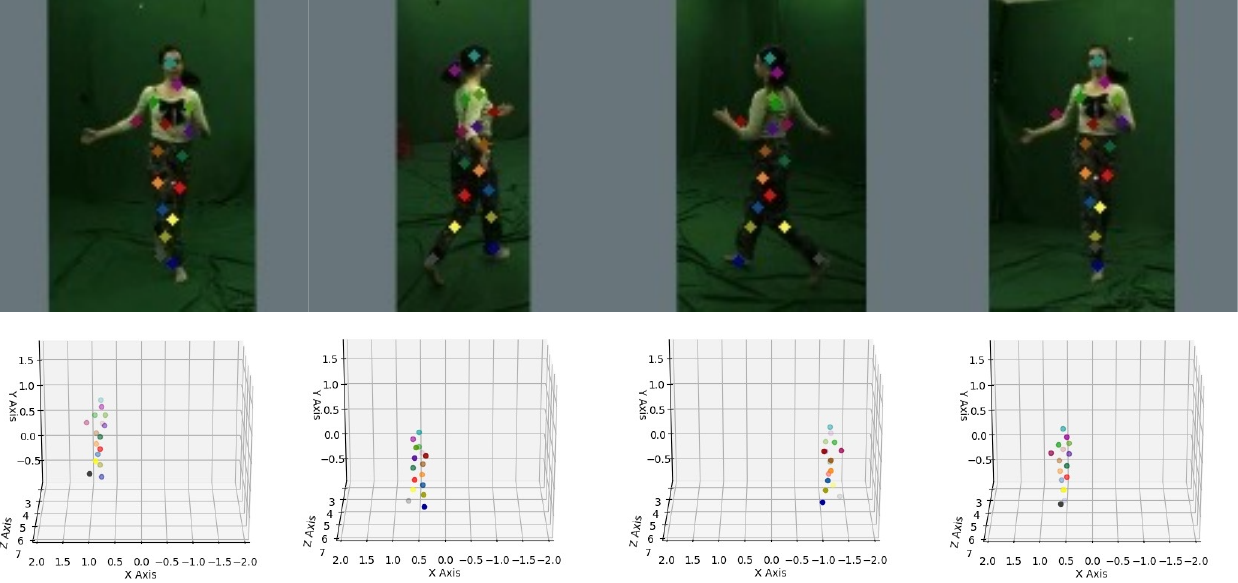}
	\end{subfigure}
	\hspace{0.5cm}
	\begin{subfigure}[b]{0.45\textwidth}
	\includegraphics[width=\textwidth]{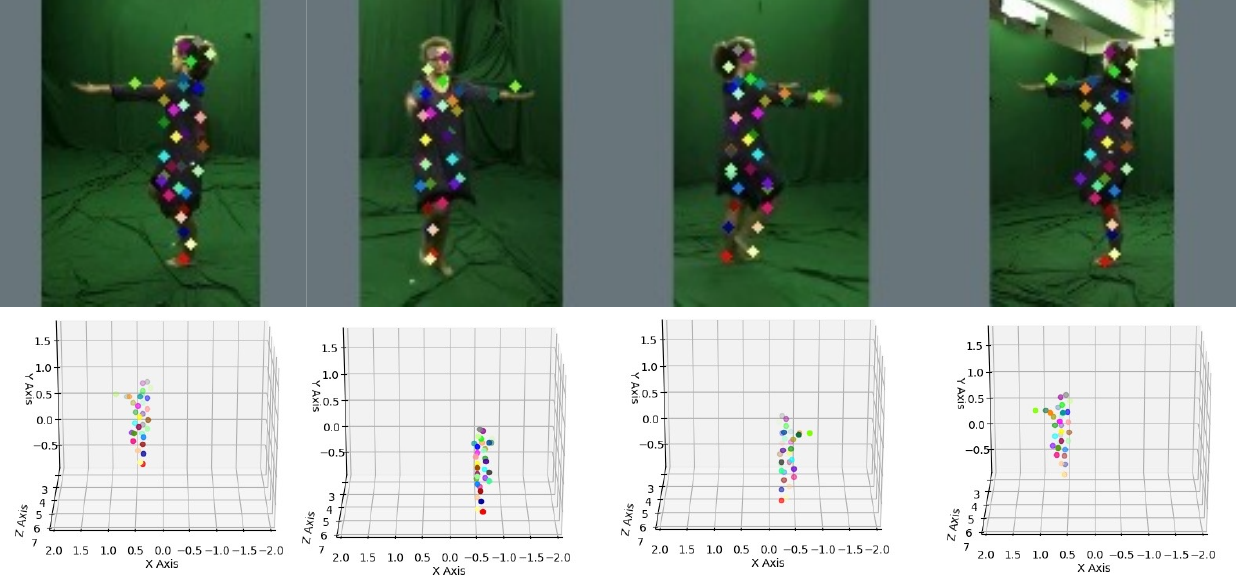}
	\end{subfigure}
	\vspace{-2mm}
	\caption{\small 
	2D and 3D keypoints found by 16- (left) and 32- (right) keypoint prediction models on 3DHP. As in Fig.~\ref{fig:kpts_3D_H36M}, the 2D keypoints are consistent across views and the 3D keypoints capture the posture of the person.}
	\label{fig:kpts_3D_3DHP}
\end{figure*}

\section{Experiments}
\label{sec:exp}

In this section, we first present the evaluation datasets and metrics. Then, we compare our model quantitatively with other unsupervised approaches and present ablation studies.

\subsection{Datasets}
\label{sec:datasets}

We use the following 3D human pose benchmark datasets.

\vspace{0.5em}
\noindent \textbf{Human3.6M \cite{Ionescu14a} (H36M).}
As in~\cite{Rhodin19a, Honari22}, we use subjects 1, 5, 6, 7, and 8 to train the unsupervised keypoint estimation model. This yields 308,760 training samples. Subjects 9 and 11 are used for evaluation, sub-sampled every 10 frames, and applied uniformly over video frames. This yields 53,720 test images.  We show qualitative results of the discovered unsupervised 2D and 3D keypoints in Fig.~\ref{fig:kpts_3D_H36M}. 

\noindent \textbf{MPI-INF-3DHP \cite{Mehta17a} (3DHP).}
We use subjects 1 to 6 for training and subjects 7 and 8 for evaluation. We take only frames where the person is the foreground subject\footnote{We leave out frames with the actor on a chair as it occludes the foreground subject.}. This yields 59,952 frames for training and 7,312 frames for evaluation. We show qualitative results of the unsupervised keypoint predictions in Fig.~\ref{fig:kpts_3D_3DHP}.

\subsection{Comparison to the State-of-The-Art Models}
\label{sec:comp}

The qualitative results of Figs.~\ref{fig:kpts_3D_H36M} and~\ref{fig:kpts_3D_3DHP} indicate that our approach captures 3D posture effectively, as the keypoints remain consistent across views. To quantify this, we compare against other unsupervised 3D and 2D approaches as well as to a pre-trained ImageNet feature extractor. We report 3D pose errors using mean per joint position error (MPJPE), the normalized N-MPJPE, and the procrustes aligned P-MPJPE between predicted and ground-truth 3D poses.

\begin{table}[ht]
	\centering
	\resizebox{1\linewidth}{!}{
            \fontsize{11pt}{11pt}\selectfont
		\begin{tabular}{p{2.7cm}p{.5cm}p{1.7cm}p{1.3cm}p{1.3cm}C{1.3cm}C{1.3cm}}
			\hline
			\textbf{Model}  &&&& \textbf{MPJPE} & \textbf{N-MPJPE} & \textbf{P-MPJPE} \\
			\hline
			\multicolumn{5}{c}{\textbf{Known Kinematic Model}} & \rule{0pt}{1\normalbaselineskip} \\
                Kundu \etal \cite{Kundu20a} &&&& 99.2 & - & - \\
			Kundu \etal \cite{Kundu20b} &&&& - & - & 89.4 \\
                \hline
                \multicolumn{5}{c}{\textbf{Uninterpretable latents}}  & \rule{0pt}{1\normalbaselineskip} \\ 
                NVS \cite{Rhodin18b} &&&& - & 115.0 & - \\
			Honari \etal \cite{Honari22} &&&& 100.3 & 99.3  & 74.9 \\
               \hline
                \multicolumn{5}{c}{\textbf{Keypoint Discovery}}  & \rule{0pt}{1\normalbaselineskip} \\
                KeypointNet \cite{Suwajanakorn18a} & SV & 2 hid MLP & 32 kpts & 158.7 & 156.8 & 112.9 \\
                Ours & SV & 2 hid MLP & 32 kpts & \textbf{125.73} & \textbf{121.04} & \textbf{89.05} \\
                \hdashline
                BKinD-3D \cite{Sun23bkind} & MV & Linear & 15 kpts\footnote{We report the best performing model with 15 kpts. It works better than the 30 kpt model.} & 125 & \rule{0pt}{1\normalbaselineskip} - & 105 \\
                Ours & MV & Linear & 32 kpts & \textbf{120.9} & \textbf{117.9} & \textbf{93.5} \\
			Ours & MV & 2 hid MLP & 32 kpts & \textbf{73.8} & \textbf{72.6} & \textbf{63.0} \\
			\hline
		\end{tabular}
	}
	\caption{\small {\bf Comparison with unsupervised 3D models on H36M (in mm)}. SV and MV indicate single- and multi-view models. kpts denotes number of keypoints.  
	}
	\label{tab:h36}
\end{table}

\begin{table}[ht]
	\centering
	\resizebox{1\linewidth}{!}{
		\begin{tabular}{lcccc}
			\hline
			\textbf{Model} & \textbf{Train-Set} & \textbf{MPJPE} & \textbf{N-MPJPE} & \textbf{P-MPJPE} \\
			\hline
			\multicolumn{4}{c}{\textbf{Uninterpretable latents}}\\ 
			DrNet~\cite{Denton17} & 3DHP & 22.28 & 21.55 & 14.94 \\
			NSD~\cite{Rhodin19a} & 3DHP &  20.24 & 19.29 & 14.09  \\
			Honari \etal~\cite{Honari22} & 3DHP & 20.95 & 19.78 & 14.04 \\ 
                \hline
                \multicolumn{4}{c}{\textbf{Keypoint Discovery}}\\
                Ours & H36M & 16.90 & 16.19 & 12.48  \\
			Ours & 3DHP & \textbf{14.57} & \textbf{14.21} & \textbf{11.52}  \\ 
			\hline
		\end{tabular}
	}
	\caption{\small 
	{\bf Comparison with unsupervised 3D models on 3DHP (in cm)}. All models use a 2 hidden layer MLP to map either features (top 3 models) or 48 discovered keypoints (Ours) to the labelled pose.
	}
	\label{tab:3dhp}
\end{table}

\parag{Unsupervised 3D Approaches.}

We compare against state-of-the-art unsupervised approaches on the H36M dataset in Table~\ref{tab:h36} and on the 3DHP dataset in Table~\ref{tab:3dhp}. We outperform the competing methods discussed in Section~\ref{sec:related}. Most importantly, our approach does not rely on any known priors such as kinematic and shape priors, and neither suffers from the lack of interpretability of the discovered features. Moreover, it outperforms other single-view and multi-view keypoint discovery approaches.

\parag{Cross Dataset Evaluation.}
We take a model trained on H36M dataset and test it on 3DHP dataset to evaluate the generalization of our approach across different datasets. The result is shown in the second to the last row in Table~\ref{tab:3dhp}. This model even outperforms other latent extraction models trained on the same test dataset.

\begin{table}[ht]
    \centering
	\resizebox{.6\linewidth}{!}{
	\begin{tabular}{|l|c|}  \hline
			\textbf{Model} & \textbf{\%-MSE Error} \\ \hline
			Thewlis et al.~\cite{thewlis19} & 7.51 \\
			Zhang et al.~\cite{Zhang18b} & 4.14 \\
			Schmidtke et al.~\cite{Schmidtke21} & 3.31 \\
			Lorenz et al. ~\cite{lorenz19} & 2.79 \\ 
			Jakab et al.~\cite{jakab20} & 2.73 \\
			Ours & \textbf{2.38} \\ \hline
		\end{tabular}
	}
	\caption{\small {\bf Comparison with 2D keypoint estimation models.} All models predict 32 keypoints on 6 actions of wait, pose, greet, direct, discuss, and walk and regress a linear model from 2D keypoints to the 2D pose labels.}
	\label{tab:h36_2D}
        \vspace{-0.1cm}
\end{table}

\parag{Unsupervised 2D models.}

For completeness, even though our primary goal is to discover 3D keypoints, we compare the discovered 2D keypoints $\bx$ of Eq.~(\ref{eq:softargmax}) with those produced by other unsupervised 2D approaches. Table~\ref{tab:h36_2D} shows that we also outperform these approaches.

\begin{figure*}[h!]
	\vspace{-0.5cm}
	\centering
	\begin{subfigure}[b]{0.25\textwidth}
		\centering
		\includegraphics[width=\textwidth]{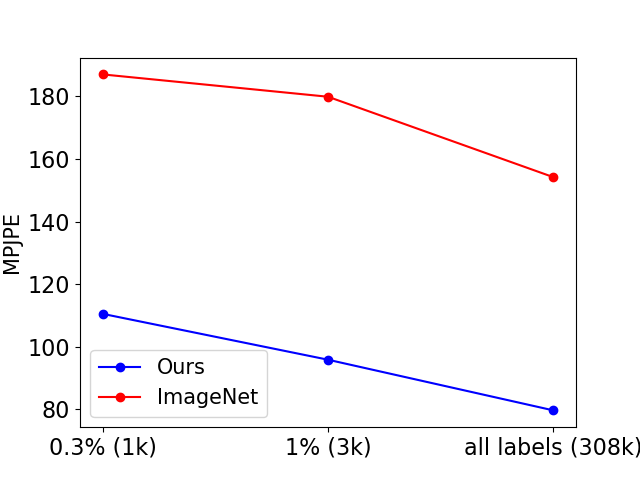}
	\end{subfigure}
	\hspace{-0.5cm}
	\begin{subfigure}[b]{0.25\textwidth}
		\includegraphics[width=\textwidth]{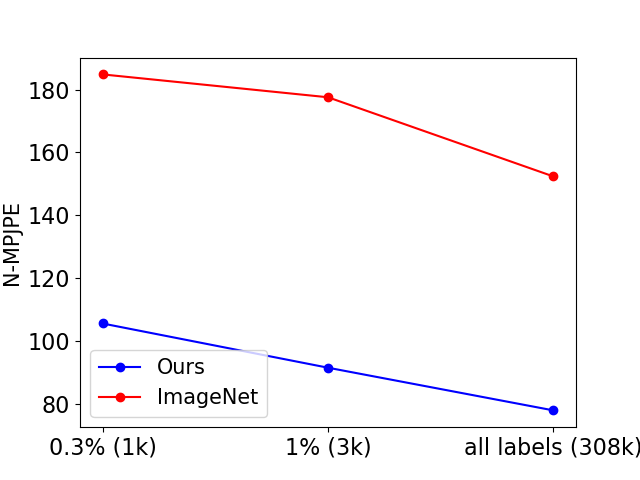}
	\end{subfigure}
	\begin{subfigure}[b]{0.25\textwidth}
	\includegraphics[width=\textwidth]{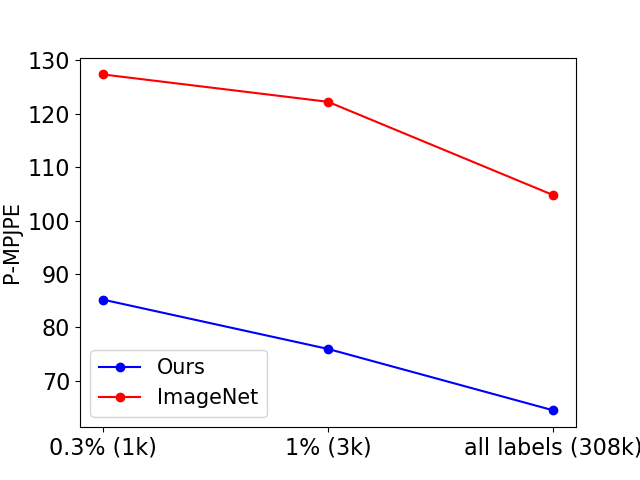}
	\end{subfigure}
	\caption{\small 
	    Comparison against the latent features of the pre-trained ImageNet model on H36M. The three plots from left to right show depict MPJPE, NMPJPE, and PMPJPE (in mm) for different percentage of labeled 3D data. Both models use a ResNet50 encoder~\cite{He16a} and leverage a 2-hidden layer MLP to regress either the 3D keypoints (Ours) or the latent features---features before the classification layer in ImageNet---to the target 3D pose.
	}
	\label{fig:imgNet_comp}
 	\vspace{-0.35cm}
\end{figure*}

\vspace{-0.1cm}
\parag{Supervised Pre-Training.}
As in recent unsupervised papers~\cite{He20c, Bao22, He22, Honari22}, we compare the quality of our unsupervised keypoints to the features obtained by a fully-supervised ImageNet model, which is commonly used for transfer learning. This evaluation highlights how much the extracted features by these models are suitable for the downstream application, which in the case of this paper is 3D human pose estimation. The results are shown in Figure~\ref{fig:imgNet_comp}. The gap in MPJPE is over 75 mm in all cases, with our approach reducing the error by about 50\%. Even in low-labeled cases, the margin is maintained. This indicates the keypoints discovered by our model are much more correlated with the target pose.

\subsection{Ablation Studies}
\label{sec:ablation}
We next evaluate the impact of loss and model components.


\subsubsection{Contribution of the Loss Components}

We first study the contribution of each loss component in Eq.~(\ref{eq:loss_unsup}). The results are presented in Table~\ref{tab:ablation_losses}. The first row only reconstructs the input image using Eq.~(\ref{eq:loss_reconst}) without any 3D keypoint estimation. In this case the latent features ${\Xi}$ are mapped to the 3D poses, as they are the closest latent representation to the pose. As observed in Table~\ref{tab:ablation_losses}, this case has about 37 mm (corresponding to 50\%) higher MPJPE error compared to our best approach in the last line. 

The second row adds the mask reconstruction loss of Eq.~(\ref{eq:loss_mask}). This improves the results by about 32 mm compared to the first row indicating the structure of the latent 3D keypoints has a considerable impact. In the third and fourth rows, we respectively add the coverage loss of Eq.~(\ref{eq:loss_coverage}) and the centering loss of Eq.~(\ref{eq:loss_bbox_center}), where the last row shows our final model with all the features activated. While these  two losses have a relatively small contribution, they still help improve the accuracy. They can be removed without hurting much the accuracy to further simplify the approach.


\begin{table}[ht]
	\centering
	  \resizebox{1\linewidth}{!}{
		\begin{tabular}{|c|c|c|c|ccc|}
			\hline
			   \textbf{$\mL_{\text{reconst}}$} &  \textbf{$\mL_{\text{mask}}$} & \textbf{$\mL_{\text{coverage}}$} & \textbf{$\mL_{\text{centering}}$} & \textbf{MPJPE} & \textbf{N-MPJPE} & \textbf{P-MPJPE} \\
			   \hline
			    \color{green}{\cmark} & \color{red}{\xmark} & \color{red}{\xmark} & \color{red}{\xmark} & 111.8 & 107.6 & 79.7 \\
			    \color{green}{\cmark} & \color{green}{\cmark} & \color{red}{\xmark} & \color{red}{\xmark} & 79.9 & 78.6 & 67.0 \\
			    \color{green}{\cmark} & \color{green}{\cmark} & \color{green}{\cmark} & \color{red}{\xmark} & 78.3 & 77.0 & 65.6 \\
			    \color{green}{\cmark} & \color{green}{\cmark} & \color{green}{\cmark} & \color{green}{\cmark} & \textbf{73.8} & \textbf{72.6} & \textbf{63.0} \\
			\hline
		\end{tabular}
	}
	\caption{\textbf{Ablation Study on Losses.} Each column shows the impact of adding a loss term in Eq.~(\ref{eq:loss_unsup}). The last row shows our model with all features activated. All models use a 2-hidden layer MLP to map the latent features (1st row) or the discovered keypoints (2nd to 4th rows) to the 3D pose labels. The results are presented in millimeters (mm) on H36M dataset, with a lower value indicating a lower error.}
	\label{tab:ablation_losses}
\end{table}


\begin{figure*}[h!]
	\vspace{-0.45cm}
	\centering
	\begin{subfigure}[b]{0.27\textwidth}
		\centering
		\includegraphics[width=\textwidth]{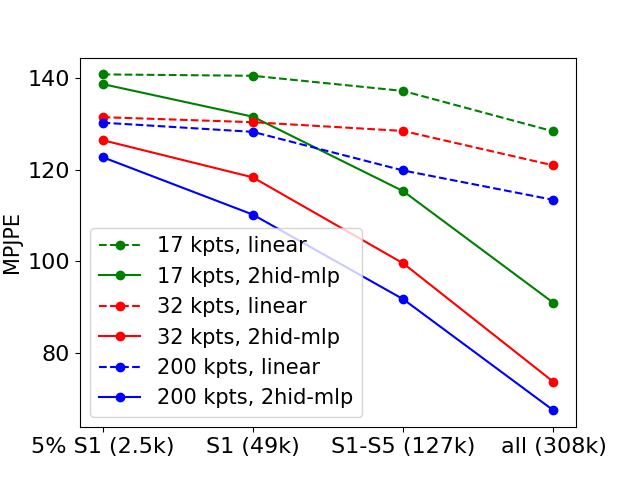}
	\end{subfigure}
	\hspace{-0.2cm}
	\begin{subfigure}[b]{0.275\textwidth}
		\includegraphics[width=\textwidth]{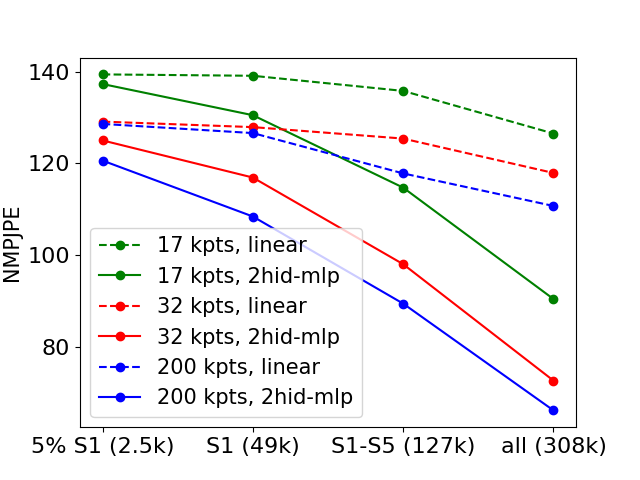}
	\end{subfigure}
	\hspace{-0.2cm}
	\begin{subfigure}[b]{0.275\textwidth}
	\includegraphics[width=\textwidth]{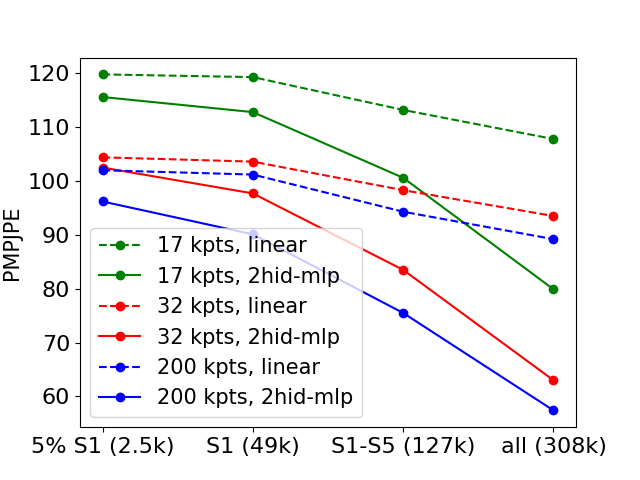}
	\end{subfigure}

	\caption{\small 
		Analysis of the impact of the number of discovered keypoints, pose model complexity, and the amount of labelled data on H36M dataset. The three plots show results on MPJPE, NMPJPE, and PMPJPE (in mm). Each plot depicts the impact of trained models using 17, 32, and 200 unsupervised keypoints. For each model, two pose-regressors are trained; one using a linear model (the dashed line) and another using a 2-hidden layer MLP (the solid line). The $X$ axes shows the percentage of 3D labels.
	}
	\label{fig:ablation_complexity}
	\vspace{-0.25cm}
\end{figure*}

\subsubsection{Keypoint, Label and Pose-Model Complexity}

We study here the impact of the size and complexity of model components, in addition to analyzing performance under different amounts of labeled data. One key question is how many unsupervised keypoints should the model predict and how much it impacts the results. Figure~\ref{fig:ablation_complexity} shows the results for 3 different numbers of keypoints:
\begin{itemize}
    \item 17: number of keypoints equal to the labeled 3D joints.
    \item 32: a number in-between the two bounds.
    \item 200: number of keypoints matching prior works~\cite{Rhodin19a, Honari22}\footnote{The dimensionality of the extracted latent features in these methods, which serves as the input dimensionality to the pose model, is 600, which equals to 200 3D keypoints.}. 
\end{itemize}

As observed in Figure~\ref{fig:ablation_complexity}, increasing the number of discovered keypoints increases accuracy. However, the gap between 17 and 32 is bigger than the one between 32 and 200 keypoints. Hence, we use 32 keypoints by default on H36M dataset to avoid adding much complexity. On 3DHP dataset we use 48 keypoints. Check Supplementary for details.

The complexity of the pose estimator model inversely correlates with the interpretability of the discovered keypoints. The more the complexity of the required pose model, the more the keypoints lie in an unrelated latent space, which requires more complex adaptation to map them to the actual pose of the subject. Our results in Figure~\ref{fig:ablation_complexity} show that while a simple 2-hidden layer MLP obtains more accurate results, a simpler linear adaptation of the keypoints to the pose labels can still obtain decent results. This further shows the robustness of the unsupervised keypoints and indicates they correlates strongly with the target pose. 

\subsubsection{Number of Views}
In order to analyze how many views are required for a robust pose estimation, we train models with 2, 3, and 4 views. The results are reported in Table~\ref{tab:ablation_views}. As expected, higher accuracy is obtained with more views, since the triangulation obtains more robust 3D keypoints. However, 3 views are enough to obtain robust results, with a clear gap with 2 views and a small gap with 4 views. For a fair comparison with previous baselines\cite{Sun23bkind,Suwajanakorn18a} that use 4 views, we use the same number of views by default in all experiments.

\subsubsection{Target Mask}
\label{sec:ablation_mask}

Our model reconstructs the mask that is estimated by the model itself in the image-reconstruction module. The quality of the obtained mask directly impacts the accuracy of the model as this is the target for the keypoint discovery components. If one has access to the ground-truth mask, how much the results would change? To answer this question we train two variants, one using the full pipeline of Section~\ref{sec:method} with all elements of Eq.~(\ref{eq:loss_unsup}), and another by omitting $\mL_{\text{reconst}}$ from it, where there is no image-reconstruction, and the encoded features $\Xi$ are used to only reconstruct the mask. The results are reported in Table~\ref{tab:ablation_mask}. While both models obtain close results, our predicted mask obtains slightly better results than the ground-truth one that is subject to noisy annotations. This highlights the robustness of our mask prediction. See examples in Supplementary.

We also compare with background reduction from the input image and  normalization in Table~\ref{tab:ablation_mask}. This approach, leads to much higher error, as the color of the subject's clothes can interfere for proper background reduction. The results indicate that while our approach does not need ground-truth masks and its estimated mask captures better the foreground subject, which leads to higher accuracy. 

\begin{table}[ht]
	\centering
	\resizebox{1\linewidth}{!}{
		\begin{tabular}{|c|c|c|c|c|}
			\hline
			\textbf{Number of Views}  & \textbf{Pose Model} &  \textbf{MPJPE} & \textbf{N-MPJPE} & \textbf{P-MPJPE} \\
			\hline
			2 & 2-hid MLP & 103.21 & 100.7 & 81.6 \\
			3 & 2-hid MLP &  77.7 & 75.7 & 64.0 \\
			4 & 2-hid MLP & \textbf{73.8} & \textbf{72.6} & \textbf{63.0} \\
			\hline
		\end{tabular}
	}
	\vspace{-0.2cm}
	\caption{\small 
	{\bf Evaluation on the number of views used for triangulation (in mm)}. In all cases 32 unsupervised 3D keypoints are predicted by the models. All models use a 2-hidden layer MLP for the pose regressor model. Results are reported on H36M dataset.
	}
	\label{tab:ablation_views}
\end{table}


\begin{table}[ht]
	\centering
	\resizebox{1\linewidth}{!}{
		\begin{tabular}{|c|c|c|c|c|}
			\hline
			\textbf{Mask Used}  & \textbf{Pose Model} &  \textbf{MPJPE} & \textbf{N-MPJPE} & \textbf{P-MPJPE} \\
			\hline
			IMG - BG & 2-hid MLP & 17.48 & 16.93 & 12.64 \\
			Ground-truth & 2-hid MLP & 14.57 & 14.21 & 11.52 \\
			Predicted (Ours)  & 2-hid MLP & \textbf{14.37} & \textbf{13.98} & \textbf{10.99} \\
			\hline
		\end{tabular}
	}
	\caption{\small 
	{\bf Impact of different target masks (in cm)}. 1st row: a mask is obtained by reducing background from the input image and normalizing it to $[0, 1]$. 2nd row: the ground-truth mask is used. 3rd row: our model's predicted mask is used to train keypoint discovery model. All models predict 48 keypoints on the 3DHP dataset and use a 2-hidden layer MLP for pose regression.
	}
	\label{tab:ablation_mask}
	\vspace{-0.2cm}
\end{table}



\section{Conclusion and Limitations}
In this paper, we proposed an unsupervised technique that leverages the full projection properties of a multi-view system to discover 3D keypoints by reconstructing the foreground subject's mask. In doing so, our model finds keypoints that are consistent across different views and correspond to the foreground subject. We show its application on real-world data featuring articulating human bodies.

There are multiple directions that our approach can be improved.
Our approach works on images with a single foreground subject. The existence of multiple foreground entities can occlude the mask, leading to extraction of keypoints that lie on all of them. Another limitation is the dependence on an estimation of the background image to help learn the mask of the foreground subject. While as we show in Section~\ref{sec:ablation_mask}, our model does not need to rely on ground-truth masks and does a better job than background subtraction from the input image, it requires a background image in the pipeline.
Moreover, our keypoints can jitter on consecutive frames of a video. Leveraging a temporal constraint can enhance the consistency of the obtained keypoints across time. We leave these extensions to future work.


\newpage
{\small
	\bibliographystyle{ieeenat_fullname}
	\bibliography{bib/string,bib/vision,bib/learning,bib/graphics,bib/egbib}
}

\newpage
\setcounter{figure}{0}

\setcounter{section}{0}
\renewcommand{\thesection}{S.\arabic{section}}
\renewcommand{\thesubsection}{\thesection.\arabic{subsection}}

\newcommand{\beginsupplementary}{%
	\setcounter{table}{0}
	\renewcommand{\thetable}{S\arabic{table}}%
	\setcounter{figure}{0}
	\setcounter{page}{1}
	\renewcommand{\thefigure}{S\arabic{figure}}%
}

\beginsupplementary
\twocolumn[{%
	\vspace{3. em}
	\centering
	\textbf{\LARGE Unsupervised 3D Keypoint Discovery with Multi-View Geometry --- Supplementary Information\\}
	\vspace{3. em}
}
]

\section{Model Architecture and Training}
In this section we describe the model architecture presented in Section~\ref{sec:method} and the training procedure.

\subsection{Unsupervised Model Architecture.}

\textbf{Detection and Image Reconstruction.}
For the image reconstruction modules describes in Section~\ref{sec:view-dependent}, the input image $\bI$ has a resolution of $500 \times 500$ pixels. The detector network $\mS$ takes the input image down-sampled by factor $4\times4$ and uses a ResNet18 \cite{He16a} architecture to output 4 parameters $(s^x, s^y, u^x, u^y)$, two specifying the scale parameters $(s^x, s^y)$ and two indicating the center of the bounding box $(u^x, u^y)$. These parameters are used to crop a patch $\bp$ from $\bI$ that is then resized to $128 \times 128$. The encoder $\mE$, which is a ResNet50~\cite{He16a} architecture, then takes the patch $\bp$ and outputs a feature map ${\Xi}$ of size 729 in H36M and 1029 in 3DHP dataset. The decoder $\bD$ takes ${\Xi}$ features and outputs a feature-map of $4 \times 128^2$, where the first three channels compose the RGB patch $\bD_{\bp}$ and the last channel yields the foreground subject mask $\bM_{\bp}$ in the patch. The architecture detail of decoder $\bD$ is presented in Table~\ref{tab:dec_arch}.

\begin{table}[ht]
	\centering
	\resizebox{1\linewidth}{!}{
		\begin{tabular}{|c|c|c|}
			\hline
			\textbf{block}  & \textbf{layers} &  \textbf{Output Dimentionality}  \\
			\hline
			  linear adaptation &  linear with dropout, ReLU &  $256 \times 16^2$ \\
			  upsampling block 1 & $2 \times 2$ bilinear-up , ($3 \times 3 $ Conv, ReLU)(3x)  & $128 \times 32^2$ \\
                upsampling block 2 & $2 \times 2$ bilinear-up, ($3 \times 3 $ Conv, ReLU)(3x) & $64 \times 64^2$ \\
                upsampling block 3 & $2 \times 2$ bilinear-up, ($3 \times 3 $ Conv, ReLU)(3x) & $32 \times 128^2$ \\
                RGB-mask output &  $1 \times 1 $ Conv & $4 \times 128^2$ \\
			\hline
		\end{tabular}
	}
	\caption{\small 
	{\bf Image decoder architecture}. 
The first block applies a fully-connected layer with dropout and ReLU with an output size of 65,536 that is reshaped to $256 \times 16^2$. These feature maps are then passed to three upsampling blocks, with each block having one bilinear upsampling of $2 \times 2$ and followed by 3 convolutional layers. Each convolutional layer has a kernel size of $3 \times 3$ together with a ReLU non-linearity. These three upsampling blocks change the feature maps from $256 \times 16^2$ respectively to $128 \times 32^2$, $64 \times 64^2$, and $32 \times 128^2$. Finally, a convolutional layer with a kernel size of $1 \times 1$ maps the features from $32 \times 128^2$ to $4 \times 128^2$ dimension, which consists of 3 RGB channels together with one mask channel.
	}
	\label{tab:dec_arch}
	\vspace{-0.2cm}
\end{table}

\textbf{Multi-View Keypoint Discovery.}
In the unsupervised keypoints path, describe in Section~\ref{sec:Unsup-3D-kpts}, the features ${\Xi}$ are passed to the keypoint encoder $\psi$, which is a network similar to the decoder $\bD$ (with distinct parameters) but with the difference that the final convolutional layer outputs features of size $N \times 128^2$, where each feature map $n \in \{1,\dots, N\}$ belongs to a different keypoint. A soft-argmax operation is applied to each channel to obtain a 2D keypoint per channel. Once 2D keypoints from all views are obtained, the triangulation of 2D to 3D keypoints and re-projection from 3D keypoints to 2D keypoints in each view are applied, as described in Section~\ref{sec:Unsup-3D-kpts}. These steps yield the re-projected keypoints $\{\hat{\bx}_n^{\vartheta}\} _{n=1}^{N}$ for each view ${\vartheta}$.

The mask decoder $\phi$ takes the set of 2D keypoins $\{\hat{\bx}_n^{\vartheta}\} _{n=1}^{N}$ and first builds a Gaussian heatmap of size $128 \times 128$ for each keypoint $\hat{\bx}_n^{\vartheta}: n \in \{1,\dots, N\}$, where the center of the Gaussian blob in the heatmap $n$ is the keypoint location $\hat{\bx}_n^{\vartheta} = (\hat{u}, \hat{v})_{n}^{\vartheta}$ and its standard deviation is set to $\sigma=0.02$. The set of $N$ Gaussian heatmaps are then concatenated together, which creates a feature map of shape $N \times 128^2$. This feature map constitutes the Gaussian heatmaps of all keypoints on the foreground subject in view ${\vartheta}$. The mask decoder then applies on these features three convolutional layers each with a kernel size of $5 \times 5$, a padding of $2 \times 2$, and a stride of $1 \times 1$. The first convolutional layer takes a feature map of size $N \times 128^2$ and outputs a feature map of size $1 \times 128^2$, with the last two convolutional layers keeping the feature map at the same size. Mask decoder's final output $\tilde{\bM}_{\bp}^{\vartheta}$ hence has a shape of $1 \times 128^2$, which can then be easily compared to the similarly sized mask $\bM_{\bp}^{\vartheta}$. Figure~\ref{fig:diagram} depicts the full architecture.

\textbf{Single-View 3D Keypoint Model.}
Regarding the model described in Section~\ref{sec:unsup_sv},
we use the architecture of \cite{Martinez17a} to lift K single-view 2D keypoints to the 3D keypoints discovered by the multi-view keypoint network. The architecture's detail is shown in Table~\ref{tab:sv_arch}. The single-view results are presented in Table~\ref{tab:h36}.


\begin{table}[ht]
	\centering
	\resizebox{1\linewidth}{!}{
		\begin{tabular}{|c|c|c|}
			\hline
			\textbf{block}  & \textbf{layers} &  \textbf{Output Dimentionality}  \\
			\hline
			  linear adaptation &  linear & 1024 \\
			  residual block 1 & linear, ReLU, linear, ReLU & 1024 \\
                residual block 2 & linear, ReLU, linear, ReLU & 1024 \\
                residual block 3 & linear, ReLU, linear, ReLU & 1024 \\
                linear adaptation &  linear & 3K \\
			\hline
		\end{tabular}
	}
	\caption{\small 
	{\bf Single-View 2D to 3D Keypoint lifting architecture}. The first layer linearly maps 2D keypoints from 2K to 1024, which is then followed by 3 residual blocks. Each residual block consists of two linear layers, each followed by a ReLU non-linearity. Each residual block adds its input to its output before passing it to the next block. The final linear layer maps latent features to 3D keypoints.
	}
	\label{tab:sv_arch}
	\vspace{-0.2cm}
\end{table}

\subsection{3D Pose Model.}
%
The 3D pose estimation model, described in Section~\ref{sec:pose_estimation}, is used to analyze the quality of the discovered unsupervised 3D keypoints. To do so, they are passed to either a linear or a 2-hidden layer MLP (depending on the evaluation setup in Section~\ref{sec:exp}), where in the latter case each hidden layer has a dimensionality of 2048 followed by a ReLU and a 50\% dropout. The output layer has a dimensionality of $3 \times M$, where $M$ is 17 labelled joints in both H36M and 3DHP datasets.

\subsection{Training Procedure.}
We train the unsupervised model's parameters for 200K iterations with mini-batches of size 32 using an Adam optimizer, which takes about 2 days on a Tesla V100 GPU. We use a learning rate of 1e-4 for all parameters, except the detector network $\mS$ and image decoder $\bD$, which use 2e-5. The pose model is also trained for 200K iterations using mini-batches of size 256 and a learning rate of 1e-4, which takes few hours on the same GPU.

Regarding the hyper-parameters of Eq.~(\ref{eq:loss_unsup}), we set $\beta$ to 2, $\delta$ to 0.01, $\eta$ to 1.0, and $\gamma$ respectively to 0.5 and 5.0 in the in H36M and 3DHP datasets.

\section{Extra Analysis}
In this section, we present further quantitative and qualitative results.

\subsection{3DHP Keypoint Impact}
In Table \ref{tab:ablation_num_kpts}, we evaluate the impact of the number of keypoints on 3DHP dataset. Similar to H36M dataset, higher accuracy is obtained with more keypoints. For this dataset, we found 48 keypoints to obtain decent accuracy without adding extra complexity. We use 48 keypoints in other evaluations done on this dataset, presented in the paper.


\begin{table}[ht]
	\centering
	\resizebox{1\linewidth}{!}{
		\begin{tabular}{|c|c|c|c|c|}
			\hline
			\textbf{Number of keypoints}  & \textbf{Pose Model} &  \textbf{MPJPE} & \textbf{N-MPJPE} & \textbf{P-MPJPE} \\
			\hline
			16 & 2-hid MLP & 16.16 & 15.76 & 12.28  \\
			32 & 2-hid MLP &  15.53 & 15.52 & 11.87  \\
			48 & 2-hid MLP & 14.57 & 14.21 & 11.52  \\
			96 & 2-hid MLP & \textbf{14.16} & \textbf{13.90} & \textbf{11.14} \\
			\hline
		\end{tabular}
	}
	\caption{\small 
	{\bf Pose accuracy given different number of discovered keypoints in 3DHP dataset (in cm)}. All models use a 2-hidden layer MLP for the pose regressor model.
	}
	\label{tab:ablation_num_kpts}
\end{table}

\subsection{Qualitative Visualization}
\begin{figure*}[h!]
    \begin{minipage}{\textwidth}
    \centering
    
    \hspace{.3cm} Image \hspace{0.6cm} Image - Bg. \hspace{0.3cm}  Mask(GT) \hspace{0.3cm} Mask(Ours)  
    \hspace{0.5cm} 
    \hspace{0.35cm} Image \hspace{0.6cm} Image - Bg. \hspace{0.3cm}  Mask(GT) \hspace{0.3cm} Mask(Ours)
    \vspace*{0.2cm}
    \end{minipage}
    \begin{minipage}{\textwidth}
    \centering
	\begin{subfigure}[b]{0.45\textwidth}
	    \includegraphics[width=\textwidth]{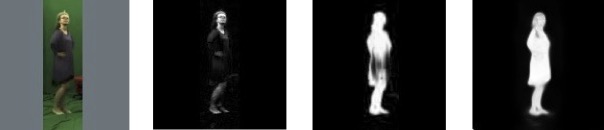}
	\end{subfigure}
	\hspace{0.5cm}
	\begin{subfigure}[b]{0.45\textwidth}
	    \includegraphics[width=\textwidth]{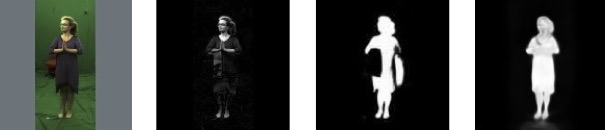}
	\end{subfigure}
	\vspace{0.2cm}
	\end{minipage}
	\begin{minipage}{\textwidth}
	\centering
	\begin{subfigure}[b]{0.45\textwidth}
		\includegraphics[width=\textwidth]{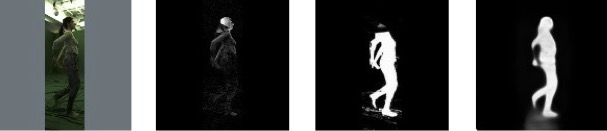}
	\end{subfigure}
	\hspace{0.5cm}
	\begin{subfigure}[b]{0.45\textwidth}
		\includegraphics[width=\textwidth]{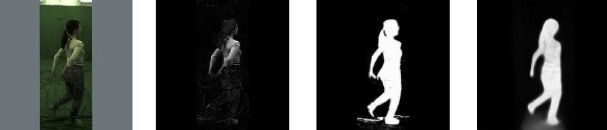}
	\end{subfigure}
	\vspace{0.2cm}
	\end{minipage}
	\begin{minipage}{\textwidth}
	\centering
	\begin{subfigure}[b]{0.45\textwidth}
        \includegraphics[width=\textwidth]{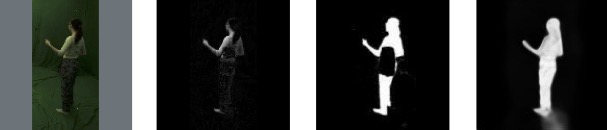}
    \end{subfigure}
    \hspace{0.5cm}
    \begin{subfigure}[b]{0.45\textwidth}
        \includegraphics[width=\textwidth]{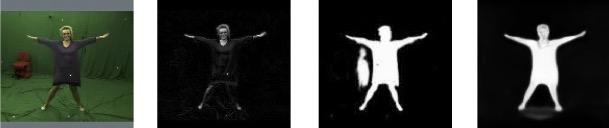}
    \end{subfigure}
    \vspace{0.2cm}
	\end{minipage}
	\begin{minipage}{\textwidth}
	\centering
    \begin{subfigure}[b]{0.45\textwidth}
        \includegraphics[width=\textwidth]{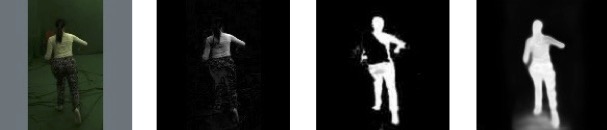}
    \end{subfigure}
    \hspace{0.5cm}
    \begin{subfigure}[b]{0.45\textwidth}
        \includegraphics[width=\textwidth]{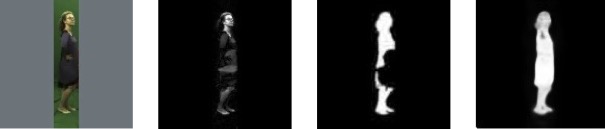}
    \end{subfigure}
    \vspace{0.2cm}
	\end{minipage}
    \begin{minipage}{\textwidth}
    \centering
    \begin{subfigure}[b]{0.45\textwidth}
        \includegraphics[width=\textwidth]{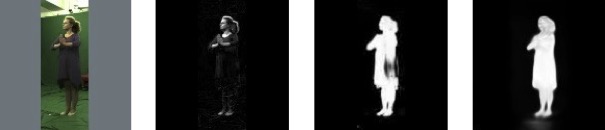}
    \end{subfigure}
    \hspace{0.5cm}
    \begin{subfigure}[b]{0.45\textwidth}
        \includegraphics[width=\textwidth]{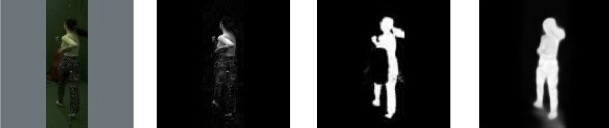}
    \end{subfigure}
    \vspace{0.2cm}
    \end{minipage}
    \begin{minipage}{\textwidth}
    \centering
    \begin{subfigure}[b]{0.45\textwidth}
        \includegraphics[width=\textwidth]{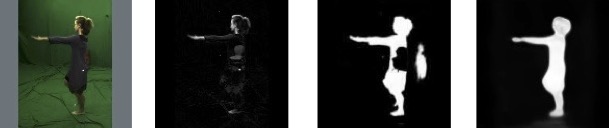}
    \end{subfigure}
    \hspace{0.5cm}
    \begin{subfigure}[b]{0.45\textwidth}
        \includegraphics[width=\textwidth]{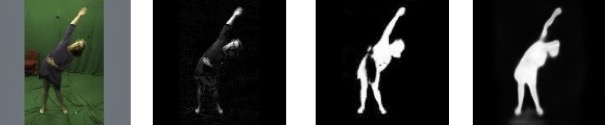}
    \end{subfigure}
    \vspace{0.2cm}
    \end{minipage}
    \begin{minipage}{\textwidth}
    \centering
    \begin{subfigure}[b]{0.45\textwidth}
        \includegraphics[width=\textwidth]{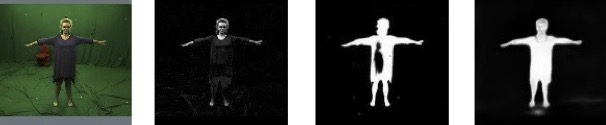}
    \end{subfigure}
    \hspace{0.5cm}
    \begin{subfigure}[b]{0.45\textwidth}
        \includegraphics[width=\textwidth]{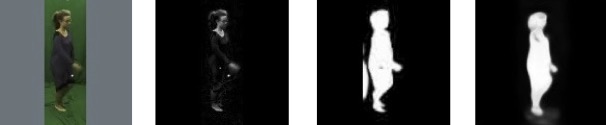}
    \end{subfigure}
    \vspace{0.2cm}
    \end{minipage}
    \begin{minipage}{\textwidth}
    \centering
    \begin{subfigure}[b]{0.45\textwidth}
        \includegraphics[width=\textwidth]{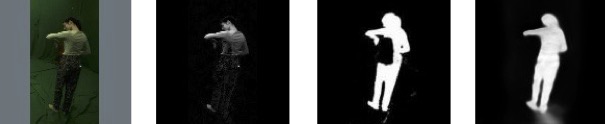}
    \end{subfigure}
    \hspace{0.5cm}
    \begin{subfigure}[b]{0.45\textwidth}
        \includegraphics[width=\textwidth]{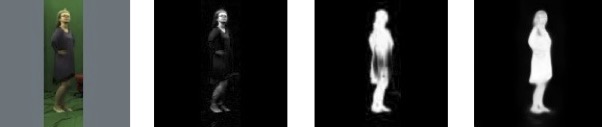}
    \end{subfigure}
    \vspace{0.2cm}
    \end{minipage}
    \rule[1ex]{17cm}{0.5pt}
    \begin{minipage}{\textwidth}
    \centering
    \begin{subfigure}[b]{0.45\textwidth}
        \includegraphics[width=\textwidth]{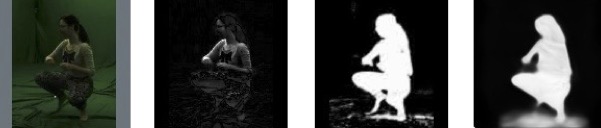}
    \end{subfigure}
    \hspace{0.5cm}
    \begin{subfigure}[b]{0.45\textwidth}
        \includegraphics[width=\textwidth]{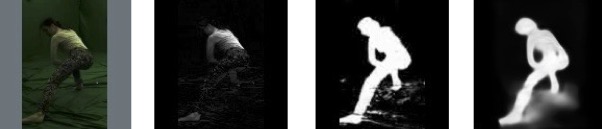}
    \end{subfigure}
    \end{minipage}
	\caption{\small \textbf{Different target masks.} For each group of 4 consecutive pictures in a row, the first column shows the input image, the second column depicts the background reduced and normalized input image, the third and fourth columns respectively show the ground truth and the mask predicted by our model. The last row shows examples where our mask creates a noticeable highlight on the floor. This is due to the shadow of the person in addition to the movement of the floor covering that the model tries to capture. Our approach learns more robust masks compared to the ground truth ones provided in the dataset that is subject to noise and inconsistent labels.}
	\label{fig:mask}
\end{figure*}

In Figure \ref{fig:mask} we show different target masks used for the experiments in Section~\ref{sec:ablation_mask}. The second columns, which depicts the mask by reducing the background from the input image and normalizing it, obviously cannot capture a mask that separates background from the subject. This is due to the similarity of the clothes of the subject with the background materials that makes the distinction not trivial. We also tried thresholding this mask with different values, but there was no clear winning threshold, as different values either removed part of the subject or included the background, which indicates a mask cannot be easily obtained in such scenarios and further highlights the importance of the prediction of the mask by the model. In the third column the ground-truth mask provided by the dataset is depicted. As can be observed, this mask also cuts parts of the subject due to noisy annotations. 

Our learned mask, on the other hand, can find the subject in a smooth and consistent way, which has a better correspondence with the subject when compared to the labels. The last row shows cases where the model also highlights the floor, which is due to the subject's shadow in addition to the change of the floor covering that the model tries to capture. While these latter examples shows the model can capture small details, the mask regions other than the foreground subject can sometimes misguide some of the keypoints, since the keypoints try to reconstruct the mask. Nevertheless, our approach still yields cleaner masks compared to the other two variants, which contributes to better performance, as shown in Table~\ref{tab:ablation_mask}. 

\begin{figure*}[h!]
	\centering
        \begin{subfigure}[b]{0.45\textwidth}
	\includegraphics[width=\textwidth]{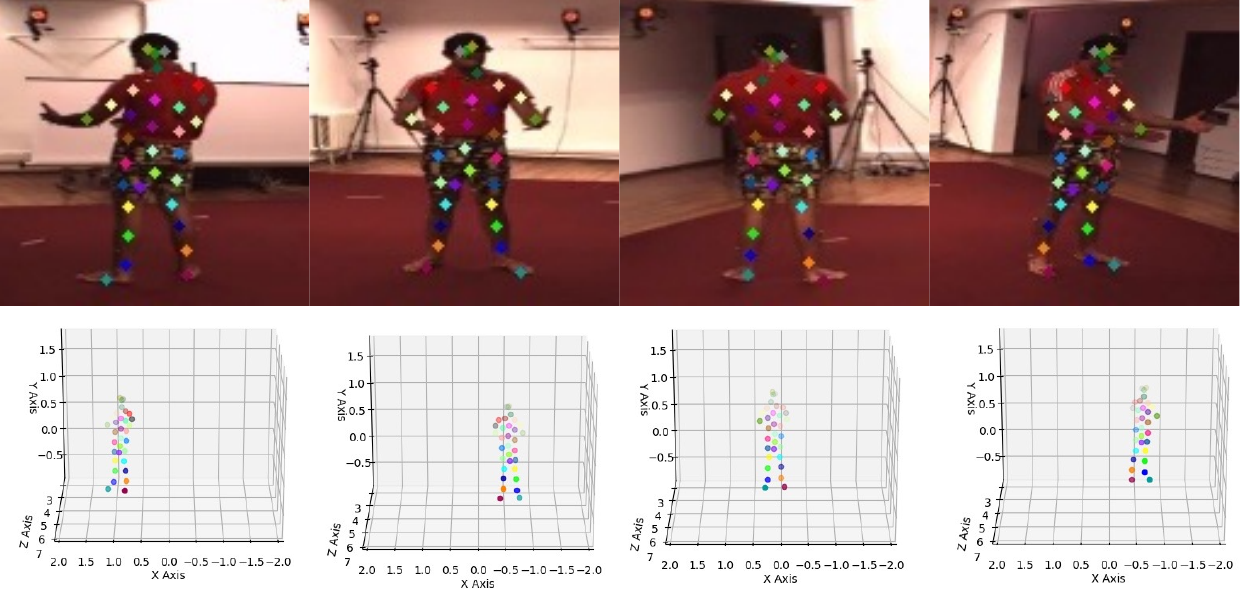}
	\end{subfigure}
	\hspace{0.5cm}
	\begin{subfigure}[b]{0.45\textwidth}
	\includegraphics[width=\textwidth]{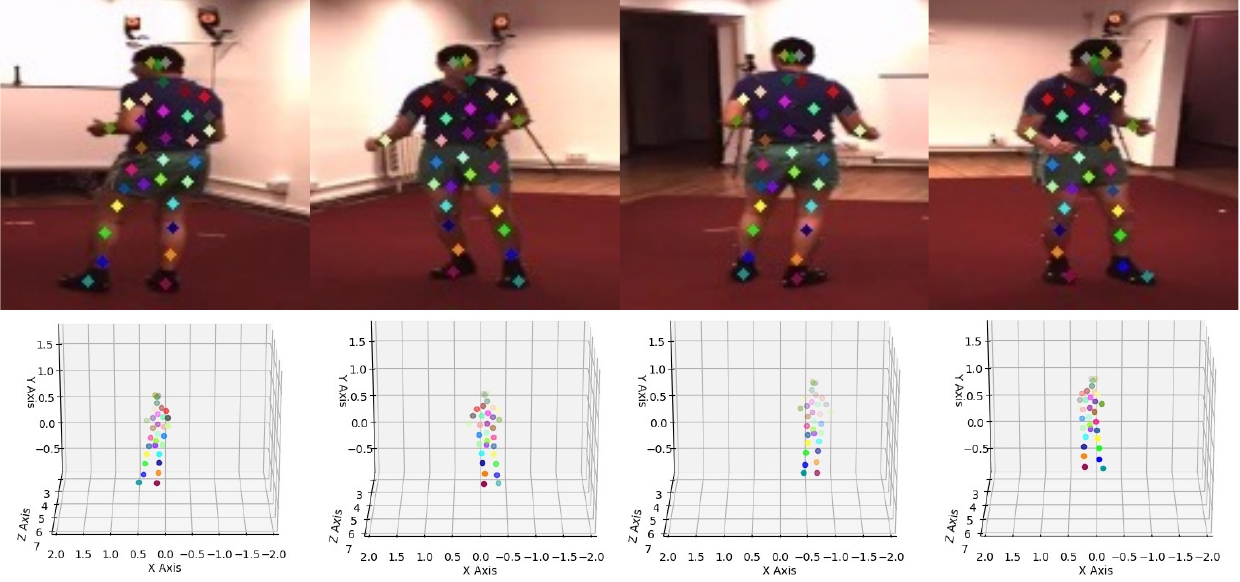}
	\end{subfigure}

        \vspace{5mm}

         \begin{subfigure}[b]{0.45\textwidth}
	\includegraphics[width=\textwidth]{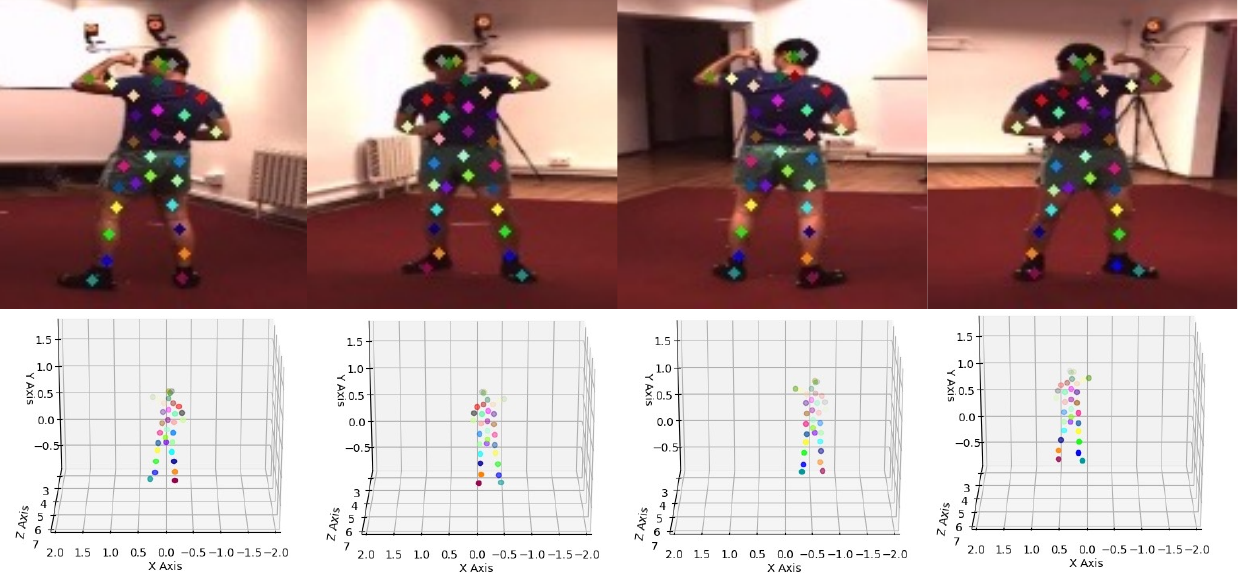}
	\end{subfigure}
	\hspace{0.5cm}
	\begin{subfigure}[b]{0.45\textwidth}
	\includegraphics[width=\textwidth]{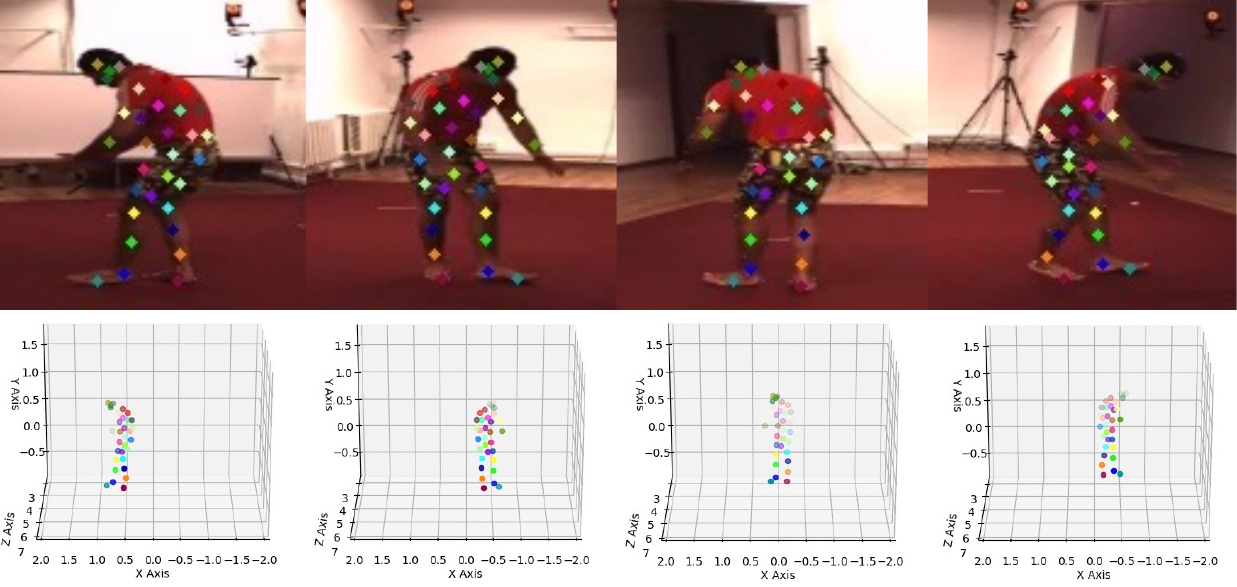}
	\end{subfigure}

        \vspace{5mm}

         \begin{subfigure}[b]{0.45\textwidth}
	\includegraphics[width=\textwidth]{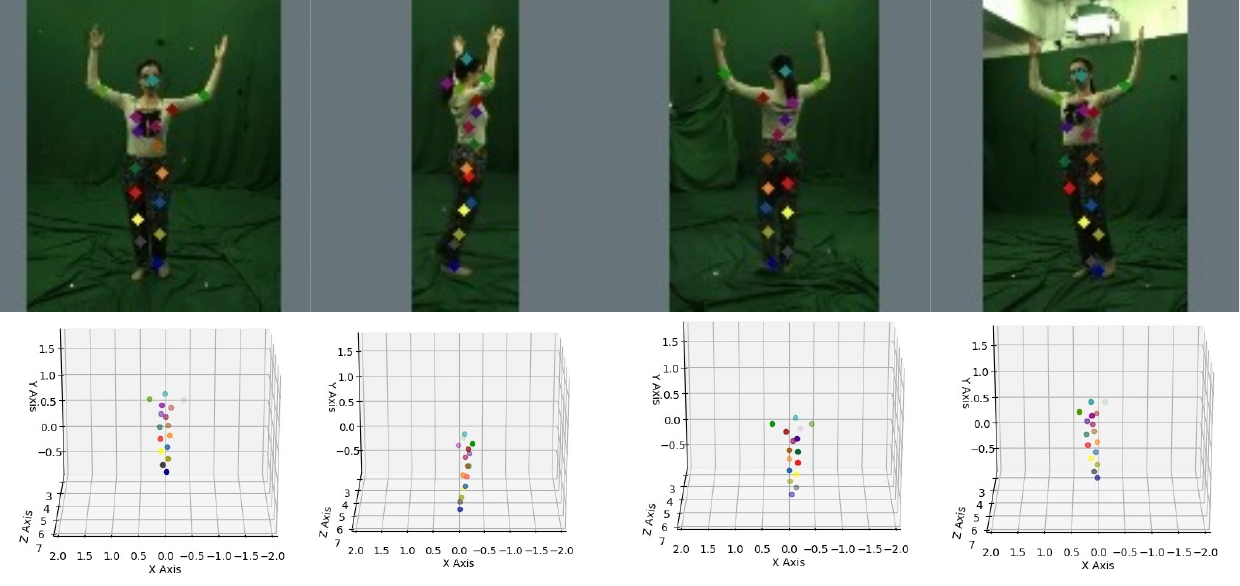}
	\end{subfigure}
	\hspace{0.5cm}
	\begin{subfigure}[b]{0.45\textwidth}
	\includegraphics[width=\textwidth]{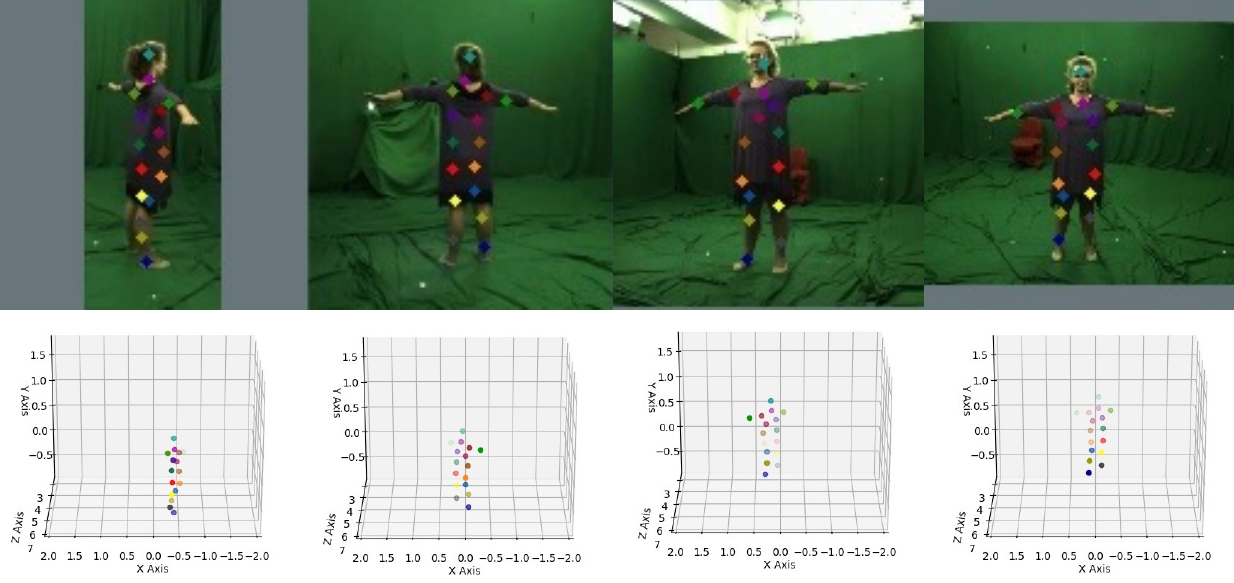}
	\end{subfigure}
 
        \vspace{5mm}

        \begin{subfigure}[b]{0.45\textwidth}
	\includegraphics[width=\textwidth]{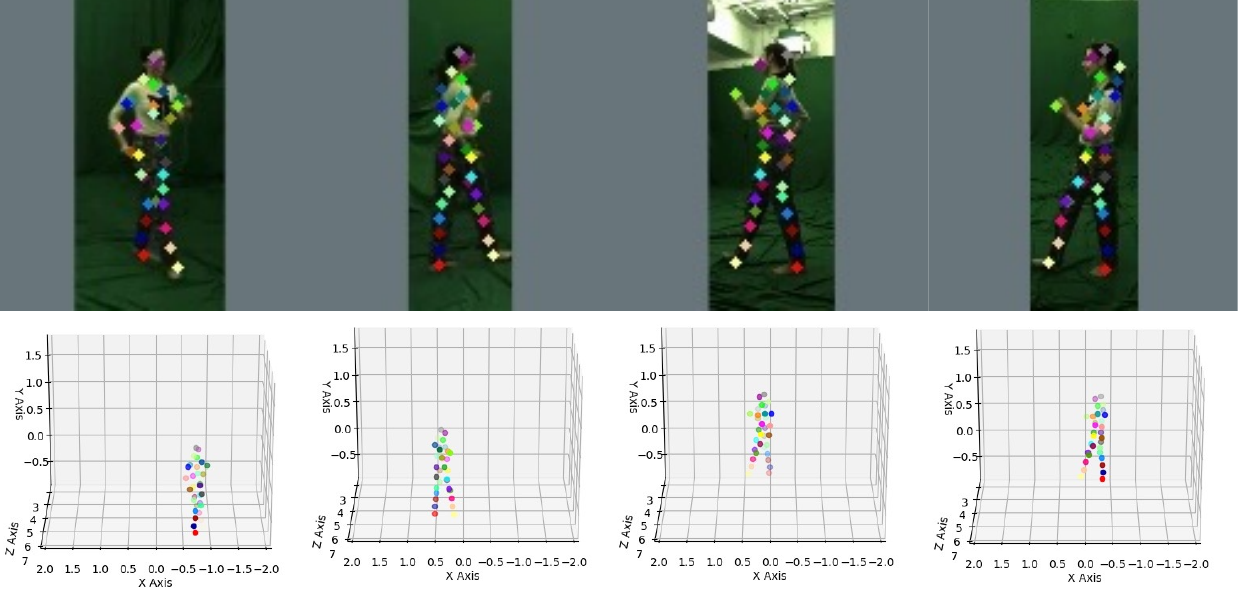}
	\end{subfigure}
	\hspace{0.5cm}
	\begin{subfigure}[b]{0.45\textwidth}
	\includegraphics[width=\textwidth]{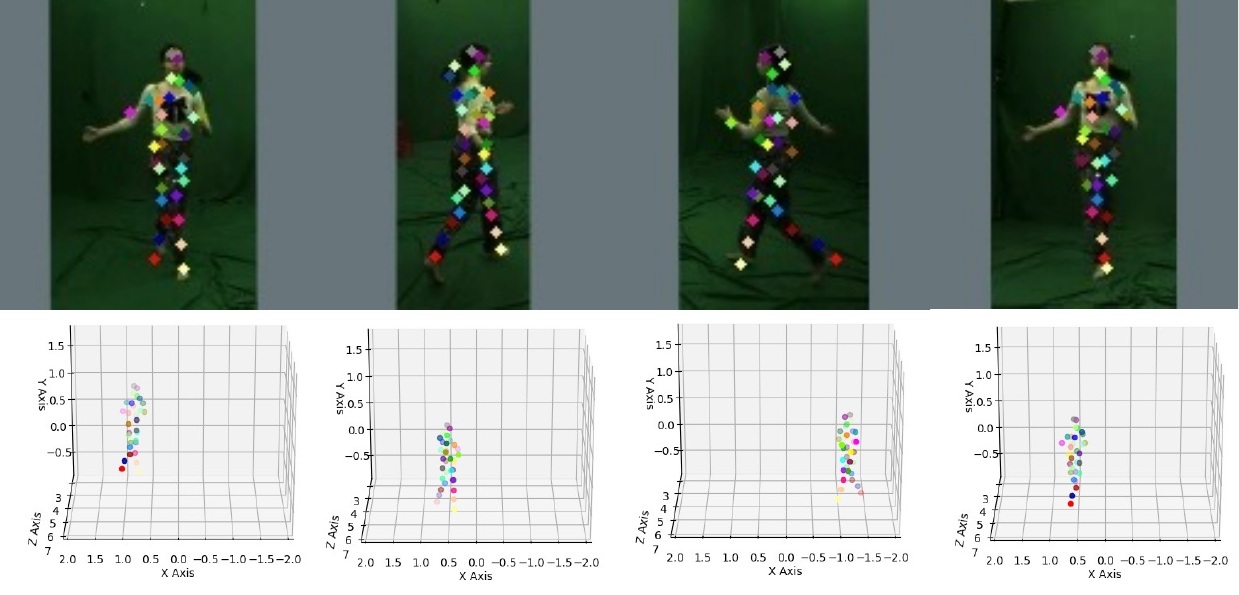}
	\end{subfigure}
  
        \vspace{5mm}

        \begin{subfigure}[b]{0.45\textwidth}
	\includegraphics[width=\textwidth]{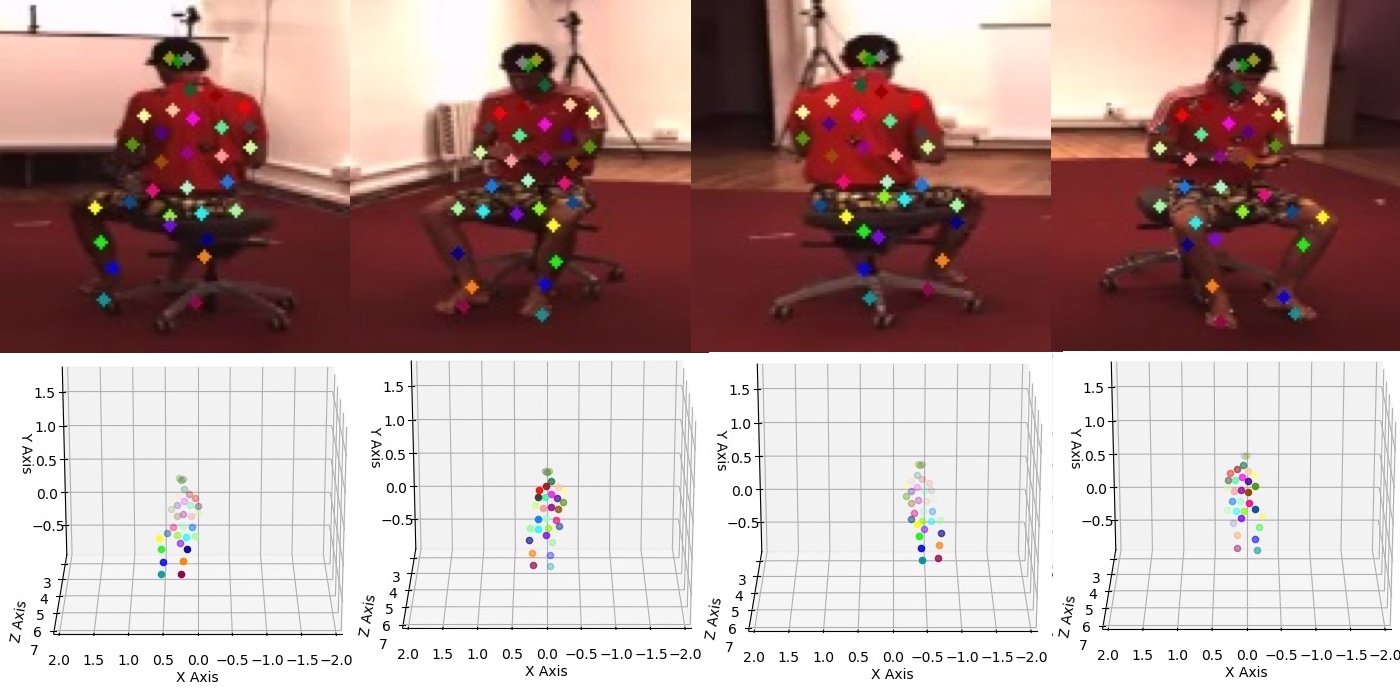}
	\end{subfigure}
	\hspace{0.5cm}
	\begin{subfigure}[b]{0.45\textwidth}
	\includegraphics[width=\textwidth]{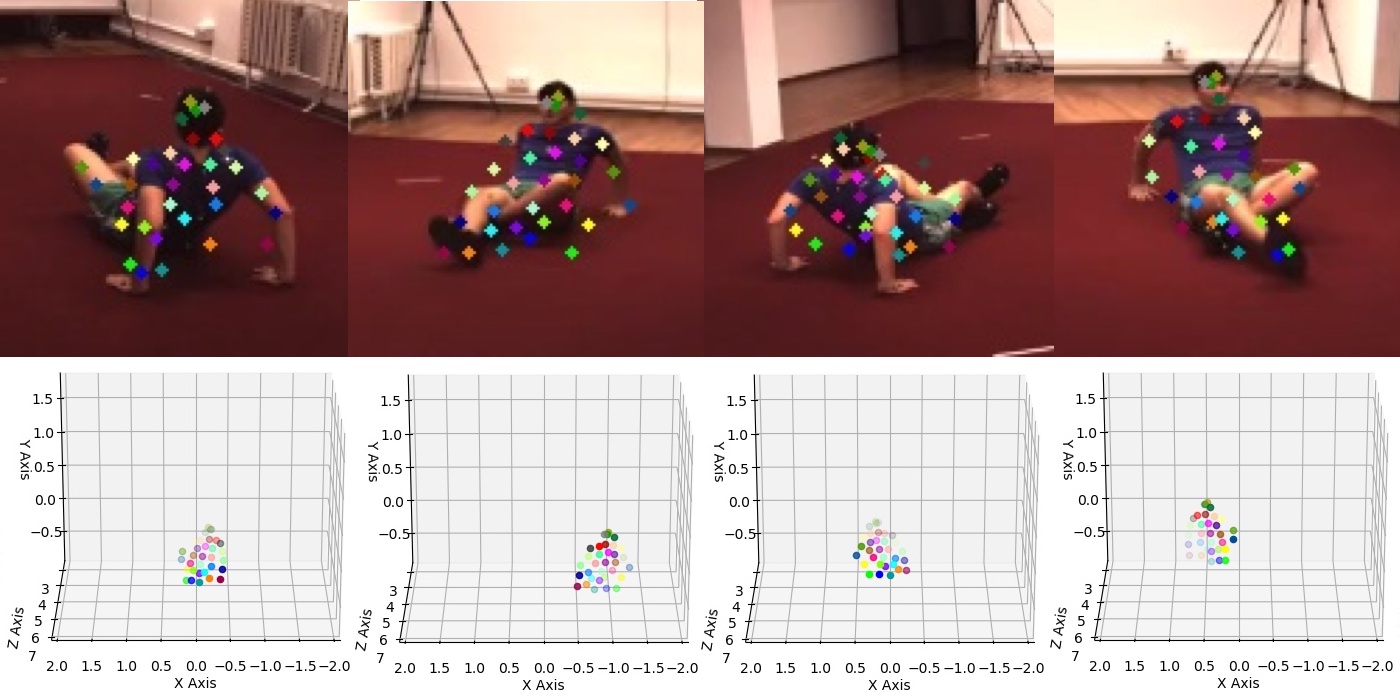}
	\end{subfigure}
 
	\caption{\small 2D and 3D keypoints discovered on H36M and 3DHP datasets. The first two rows show 32 discovered keypoints on the H36M dataset. The third and fourth rows show respectively 16 and 32 keypoints found on the 3DHP dataset. The last row shows erroneous cases, where in the left example the model finds keypoints on the chair, which is the other foreground subject. In the right example on the bottom row, the model finds keypoints on the floor due to shadow of the person. These example show cases where existence of other objects or foreground hallucination due to shadow can use errors in both the foreground mask and also the detected keypoints.}
	\label{fig:kpts_3D_supp}
\end{figure*} 
In Figure \ref{fig:kpts_3D_supp} we provide further visualization of 2D and 3D keypoints discovered by our model.
We also provide supplementary videos showing the 2D and 3D keypoints predicted by our model on H36M and 3DHP datasets. We provide 4 videos; two showing results on H36M with 17 and 32 keypoints, and the other two showing results on 3DHP with 16 and 32 keypoints. In each video, the top row depicts 2D keypoints and the bottom row shows the 3D keypoints (in camera coordinate) from different views. The keypoints are color-coded and the same color corresponds to the same keypoint in all cameras.

The videos show that the 2D and 3D keypoints capture the posture of the person, hence they correlate strongly with the actual pose of the person. We observe that when the number of keypoints is low, the model can miss parts of the subject such as hands. On the hand, when the number of keypoints is high the model can capture details other than the subject such as the subject's shadow or the changing background. Indeed, the correct number of keypoints should to be chosen to find the right balance. Finally, we observe that the keypoints can jitter across videos. We believe addition of a temporal consistency component can further help in stabilizing the keypoints. 

\end{document}